\documentclass[journal]{IEEEtran}
\usepackage{times}
\usepackage{epsfig}
\usepackage{graphicx}
\usepackage{amsmath}
\usepackage{amssymb}
\usepackage{stfloats}
\usepackage{threeparttable}
\usepackage{booktabs}
\usepackage{multirow}
\usepackage{url}
\usepackage{xcolor}
\usepackage{colortbl}
\usepackage{subcaption}
\usepackage{makecell}
\usepackage[pagebackref=true,breaklinks=true,colorlinks,bookmarks=false]{hyperref}
\usepackage{color}

\usepackage{listings}
\definecolor{codegreen}{rgb}{0,0.6,0}
\definecolor{codegray}{rgb}{0.5,0.5,0.5}
\definecolor{codepurple}{rgb}{0.58,0,0.82}
\definecolor{backcolour}{rgb}{0.95,0.95,0.92}
\lstdefinestyle{mystyle}{
	backgroundcolor=\color{backcolour},   commentstyle=\color{codegreen},
	keywordstyle=\color{magenta},
	numberstyle=\tiny\color{codegray},
	stringstyle=\color{codepurple},
	basicstyle=\ttfamily\footnotesize,
	breakatwhitespace=false,         
	breaklines=true,                 
	captionpos=b,                    
	keepspaces=true,                 
	numbers=none,                    
	numbersep=5pt,                  
	showspaces=false,                
	showstringspaces=false,
	showtabs=false,                  
	tabsize=2
}
\usepackage{graphicx}
\newcommand{\etal}{et al. }
\newcommand{\ie}{i.e., }
\newcommand{\eg}{e.g., }
\newcommand{\etc}{etc. }
\usepackage{algorithm}
\usepackage{algorithmic}

\ifCLASSINFOpdf
\else
\fi

\begin{document}

\title{Structured Attention Composition for\\Temporal Action Localization}
\author{Le Yang, Junwei Han, \IEEEmembership{Fellow,~IEEE,} Tao Zhao, Nian Liu, Dingwen~Zhang~\IEEEmembership{Member,~IEEE}
\thanks{This work was supported in part by the National Key R\&D Program of China under Grant 2021B0101200001; the Key R\&D Program of Shaanxi Province under Grant 2021ZDLGY01-08; the National Natural Science Foundation of China under Grant  U21B2048, 62136007, 62036011, and 6202781, and Zhejiang Lab (No.2019KD0AD01/010). (\emph{Corresponding authors: Junwei Han and Dingwen Zhang.})}
\thanks{L. Yang, J. Han, T. Zhao and D. Zhang are with the Brain Lab (\url{https://nwpu-brainlab.gitee.io/index_en.html}), Northwestern Polytechnical University. N. Liu is with the Inception Institute of Artificial Intelligence, Abu Dhabi, UAE. (e-mails: junweihan2010@gmail.com and zhangdingwen2006yyy@gmail.com.)}
}

\markboth{IEEE TRANSACTIONS ON IMAGE PROCESSING}%
{Shell \MakeLowercase{\textit{et al.}}: Bare Demo of IEEEtran.cls for IEEE Journals}

\maketitle

\begin{abstract}
Temporal action localization aims at localizing action instances from untrimmed videos. Existing works have designed various effective modules to precisely localize action instances based on appearance and motion features. However, by treating these two kinds of features with equal importance, previous works cannot take full advantage of each modality feature, making the learned model still sub-optimal. To tackle this issue, we make an early effort to study temporal action localization from the perspective of multi-modality feature learning, based on the observation that different actions exhibit specific preferences to appearance or motion modality. Specifically, we build a novel structured attention composition module. Unlike conventional attention, the proposed module would not infer frame attention and modality attention independently. Instead, by casting the relationship between the modality attention and the frame attention as an attention assignment process, the structured attention composition module learns to encode the frame-modality structure and uses it to regularize the inferred frame attention and modality attention, respectively, upon the optimal transport theory. The final frame-modality attention is obtained by the composition of the two individual attentions. The proposed structured attention composition module can be deployed as a plug-and-play module into existing action localization frameworks. Extensive experiments on two widely used benchmarks show that the proposed structured attention composition consistently improves four state-of-the-art temporal action localization methods and builds new state-of-the-art performance on THUMOS14.
\end{abstract}


\IEEEpeerreviewmaketitle

\section{Introduction}
\IEEEPARstart{T}{emporal} action localization is a fundamental step towards intelligent video understanding\cite{lin2020key, zhang2021weakly}. Given an untrimmed video, this task aims at localizing each action instance via predicting the start time, end time, and action category. Starting from the two-stream network \cite{wang2016temporal}, researchers \cite{carreira2017quo, wang2018temporal} have verified the efficacy and necessity of simultaneously utilizing appearance features and motion features. As for current action localization research, some works \cite{li2020deep, xu2020g, zhao2020bottom, su2020bsnPP, lin2019bmn} equally treat these two feature modalities and concatenate two kinds of features, where the complementarity of the two modalities is mined by the subsequent network in a data-driven manner. Other works \cite{ liu2021multi, lin2021learning, zeng2019graph} first learn two action localization models from two feature modalities separately, and then fuse the localization results via one trade-off coefficient.

\begin{figure*}[t]
	\graphicspath{{figure/}}
	\centering
	\includegraphics[width=1\linewidth]{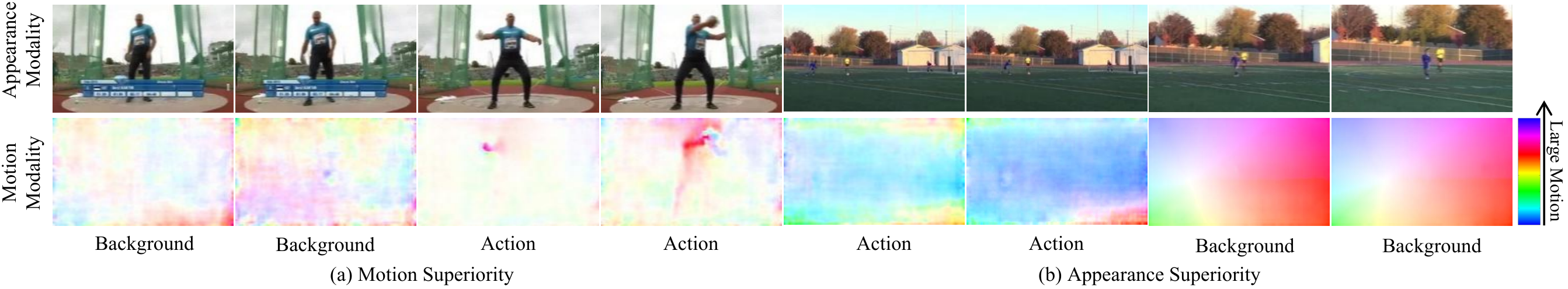}
	\caption{(a) An example of motion superiority. The appearances before and after the starting of \textit{Throw Discus} is similar, but the variance in motion modality clearly indicates the start moment. (b) An example of appearance superiority. The far camera view makes the motion modality unclear and would confuse the localization process, while the appearance modality along with context cues (\ie the football goal) can indicate an action instance. The legend indicates the motion intensity of each pixel.}
	\label{fig-motivation_RGB_flow_preference}
	\vspace{-0.5cm}
\end{figure*}

Although temporal action localization has been adequately developed, the differences between appearance modality and motion modality are inadvertently neglected, making the learned model still sub-optimal. Recently, video recognition research \cite{wang2020makes} has noticed that integrating appearance features and motion features would increase or decrease the performance, depending on whether an effective integration strategy can be designed. As shown in Fig. \ref{fig-motivation_RGB_flow_preference}, appearance modality and motion modality exhibit different superiorities to action instances. However, insufficient modeling about the impact of each modality and each frame limits the performance of existing works.

To alleviate this issue, we make an early effort to localize actions from the perspective of multi-modality feature learning. Intuitively, once we dynamically infer the frame-modality attention weight, the related frame and modality should respond loudly in the action localization process. To this end, a simple choice is to infer the frame-modality attention via convolutional layers. Nevertheless, it is insufficient, because only local features are utilized, and more importantly, the structure information (\ie modality-frame relationships) cannot be explored. Besides, there exist other choices to capture both local and global information, \eg non-local \cite{wang2018non}, or sequentially calculating modality attention and frame attention. However, these choices independently predict the attention weight for each modality and each frame, and do not constrain the mutual relationship among multiple attention weights, \ie missing the structure information.

This paper builds a novel structured attention composition module to perform multi-modality feature learning for temporal action localization. Specifically, we first separately infer the modality attention and frame attention, where the former utilizes global features to judge the modality superiority and the latter estimates potential action frames from local features. Then, we compose the frame-modality attention. Moreover, to capture the frame-modality structure, we cast the learning between the modality attention and the frame attention as a \emph{structure-guided attention assignment} process. In detail, we regard the modality attention as the supplier and the frame attention as the receiver, where each assignment from supplier to receiver would cause a particular cost. The assignment process aims at figuring out an optimal assignment plan to minimize the holistic assignment cost. Based on the optimal transport theory \cite{cuturi2013sinkhorn}, such a process could justify the fitness between the suppliers and the receivers by considering the close or alien relationship between each supplier-receiver pair. Unlike previous works \cite{ge2021ota, maretic2019got, chen2020graph}, because there is no prior knowledge about the cost matrix, we utilize two modality features to estimate the cost matrix, which serves as the frame-modality structure matrix. In summary, the proposed structured attention composition module learns the frame-modality structure matrix, and uses it to justify the fitness between the inferred modality attention and frame attention.

The framework of the proposed structured attention composition module is shown in Fig. \ref{figFramework}. We employ appearance features and motion features to predict the frame attention, modality attention and structure matrix. Then, the optimal transport process justifies the fitness between modality attention and frame attention, and learns the frame-modality structure. To assist the estimation of the structure matrix, we introduce the smoothness regularizer. Besides, an action-aware pooling strategy is proposed to estimate the modality attention. In summary, our contributions are three-fold:
\begin{itemize}
	\item 
	We make an early effort to study temporal action localization from the perspective of multi-modality feature learning, which is neglected by previous works but has an important influence on localization performance.
	\item 
	We propose the structured attention composition module to precisely infer frame-modality attention. Under the optimal transport process, the structure information is obtained by the learnable matrix, which can justify the fitness between frame attention and modality attention.
	\item 
	The proposed structured attention composition module can serve as a plug-and-play module for existing temporal action localization methods. We introduce it into previous methods \cite{xu2020g, liu2021multi, lin2017single, lin2019bmn} and consistently receive performance gains, \eg based on \cite{liu2021multi}, we achieve new state-of-the-art performance, \ie 57.6\% mAP on the THUMOS14 dataset under threshold 0.5.
\end{itemize}

\section{Related work}

\textbf{Temporal action localization} aims at discovering action instances from untrimmed videos. Pioneering works employ hand-crafted features, \eg the pyramid of score distribution \cite{yuan2016temporal}. Then, the convolutional network brings representative features to video recognition \cite{tran2015learning, wang2016temporal, carreira2017quo} and action localization tasks. For example, R-C3D \cite{xu2017r} employs the C3D \cite{tran2015learning} network to capture spatio-temporal features. Later, in addition to appearance modality, researchers \cite{gao2017turn, zhao2020temporal} notice the effectiveness of motion modality. Gao \etal \cite{gao2017turn} demonstrate that the motion modality performs superiorly to the appearance modality, and Lin \etal \cite{lin2017single} and Zhao \etal \cite{zhao2020temporal} analyze the influences of each modality in detail. Gradually, it becomes a consensus to utilize appearance modality and motion modality simultaneously \cite{gao2017cascaded,  liu2019multi, 2020Revisiting, liu2021weakly}. Most current works focus on improving action localization performance via designing effective modules, \eg anchor-based head \cite{lin2017single, xu2017r,  guo2021strengthen}, anchor-free head \cite{2020Revisiting, lin2021learning, hsieh2022contextual, huang2021clrnet, zhu2021learning}, graph convolutional network \cite{zeng2019graph, xu2020g}, refinement module \cite{liu2020progressive, qing2021temporal, xia2022learning}, and gaussian modeling \cite{long2019gaussian, zhao2021video}, but leave the role of modality under-explored. Unlike current works, this paper makes an early effort to study temporal action localization from the perspective of multi-modality learning, and develops an attention assignment process to capture the structure information. It should be noticed that this work focuses on the temporal action localization task, and leaves other related tasks for future explorations. For example, the high-quality, large-scale HiEve \cite{lin2020human} dataset for video analysis, the CFAD \cite{li2020cfad} framework performing spatio-temporal action detection in a coarse-to-fine manner.

\textbf{Multi-modality learning} strives to integrate the common information across multiple modalities, while keeps the specific pattern within each one \cite{baltruvsaitis2018multimodal, wang2020deep}. The aggregation-based paradigm \cite{zeng2019deep, valada2019self} first independently tackles each modality input, and then aggregates high-level representations, while the alignment-based paradigm \cite{cheng2017locality, song2020modality} aligns the intermediate representation of each modality. As an early effort to perform multi-modality learning on temporal action localization, one distinguishable characteristic of the proposed structured attention composition module is that it casts the frame-modality attention estimation as an attention assignment problem and solves it via the optimal transport theory.

\textbf{Structured attention} mechanism develops traditional soft-selection attention mechanism via incorporating the structure information. Kim \etal \cite{kim2017structured} utilize the graphical model to introduce structural distributions into deep networks, and receive gains on natural language processing. Later, Xu \etal \cite{xu2018structured} utilize CRF to estimate the attention map and constrain neighboring pixels' attention weights to be related. In contrast, this paper tackles the different temporal action localization tasks and utilizes optimal transport to justify the fitness between the modality's total attention and each frame's attention, developing action localization from a novel multi-modality learning perspective.

\textbf{Optimal transport theory} is preliminarily developed for measuring the geometric distance between two probability distributions. The Sinkhorn-Knopp iteration \cite{cuturi2013sinkhorn} reduces the computational complexity via introducing the entropy-regularization term. Afterward, the optimal transport theory demonstrates its efficacy in multiple domains, \eg graph matching \cite{maretic2019got, Puy2020FLOT}, cross-domain alignment \cite{chen2020graph}, object detection \cite{ge2021ota, zhang2020weakly, zhang2020discriminant, huang2022robust}. This work introduces the optimal transport process to guide the learning of modality attention and frame attention. Besides, different from previous works that utilize fixed or designed cost matrices, we propose to learn a structure matrix to capture structure information among modalities and each frame.

\section{Method}

\begin{figure*}[t]
	\graphicspath{{figure/}}
	\centering
	\includegraphics[width=0.9\linewidth]{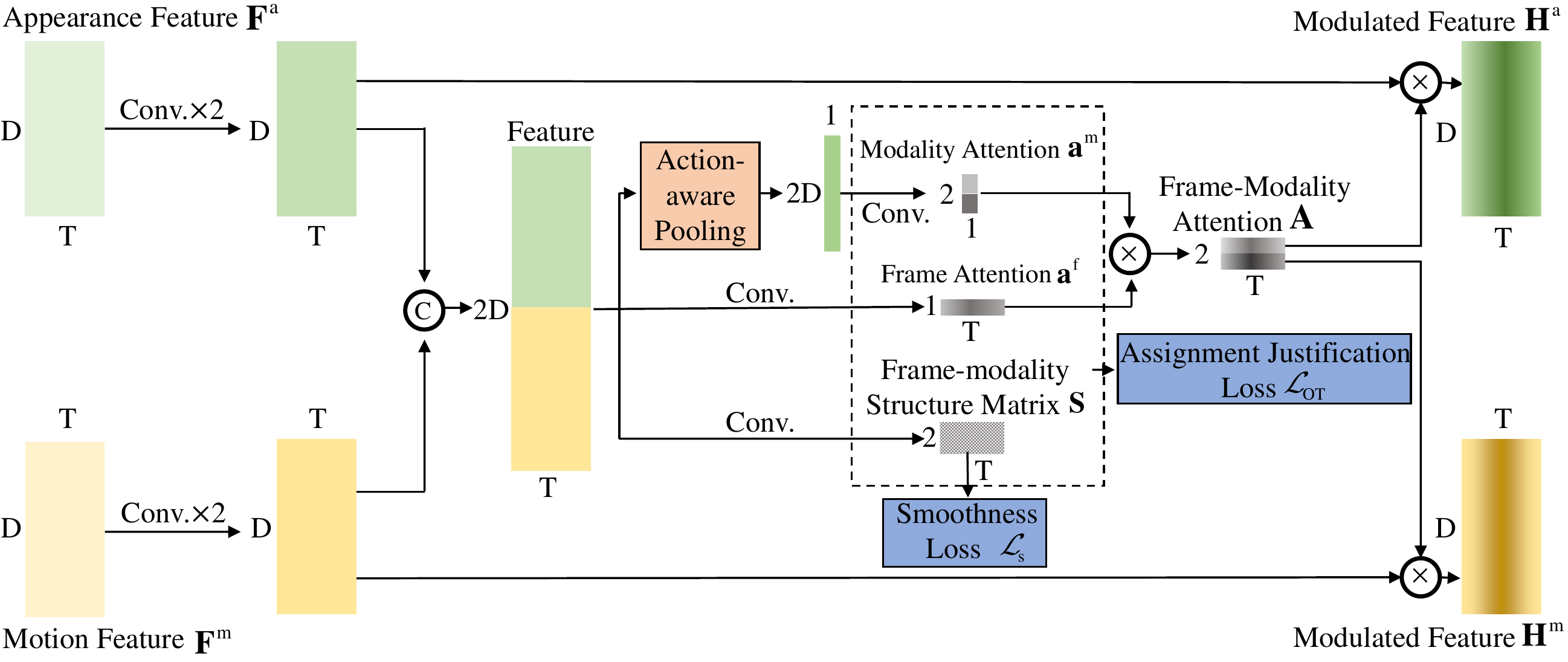}
	\caption{Framework of the proposed structured attention composition module. Given appearance features and motion features, we infer modality attention and frame attention, respectively, and employ them to compose frame-modality attention. In particular, we cast the attention estimation process as an attention assignment problem, evaluate the assignment justification via the optimal transport theory, and capture the frame-modality structure via a learnable structure matrix. The structured attention composition module can serve as a plug-and-play module, modulate vanilla appearance features and motion features, and consistently improves existing temporal action localization methods.}
	\label{figFramework}
	\vspace{-0.5cm}
\end{figure*}

Given an untrimmed video, we can extract features from the appearance modality $\mathbf{F}^{\rm a}=[\mathbf{f}^{\rm a}_{1}, \mathbf{f}^{\rm a}_{2}, ..., \mathbf{f}^{\rm a}_{T}]$ and the motion modality $\mathbf{F}^{\rm m}=[\mathbf{f}^{\rm m}_{1}, \mathbf{f}^{\rm m}_{2}, ..., \mathbf{f}^{\rm m}_{T}]$. Here, $\mathbf{F}^{\rm a}, \mathbf{F}^{\rm m} \in \mathbb{R}^{D \times T}$, where $D$ indicates feature dimension and $T$ indicates temporal length of the feature sequence. In the traditional paradigm, most works \cite{li2020deep, xu2020g, liu2021multi, lin2021learning} strive to improve the localization performance via elaborate designs but leave the role of modality under-explored. In this work, we make an early effort to study temporal action localization from the perspective of multi-modality learning, and propose a plug-and-play structured attention composition module. As shown in Fig. \ref{figFramework}, we aim at estimating the frame-modality attention $\mathbf{A} \in \mathbb{R}^{2 \times T}$, and use it to explicitly distinguish the influence of each modality and each frame, \eg $\mathbf{H}^{\rm a} = \mathbf{F}^{\rm a} \cdot \mathbf{A}[1, :]$, $\mathbf{H}^{\rm m} = \mathbf{F}^{\rm m} \cdot \mathbf{A}[2, :]$. In this way, the structured attention composition module can be introduced to existing temporal action localization methods and help localize action instances from multi-modality learning.

\subsection{Attention composition}
To infer the frame-modality attention $\mathbf{A}$, intuitively, we can first merge features from two modalities, \eg via concatenation, then predict frame-modality attention via convolutional layers. However, such estimation only captures local information within neighboring frames, which cannot thoroughly explore the effectiveness of the frame-modality attention.

We propose the attention composition strategy to utilize local cues and global cues simultaneously. Specifically, we predict the frame attention $\mathbf{a}^{\rm f} \in \mathbb{R}^{1 \times T}$ via a convolutional layer followed by the \textit{Sigmoid} activation, and predict the modality attention $\mathbf{a}^{\rm m} \in \mathbb{R}^{2 \times 1}$ via a convolutional layer followed by the \textit{Softmax} activation. The frame attention can distinguish action frames from surrounding backgrounds and captures local information. The modality attention sets superiority to the appearance modality or the motion modality, and captures global information. Given frame attention $\mathbf{a}^{\rm f}$ and modality attention $\mathbf{a}^{\rm m}$, we can compose the frame-modality attention, \ie $\mathbf{A} = \mathbf{a}^{\rm m} \times \mathbf{a}^{\rm f}$. Compared with jointly estimating the frame-modality attention, the attention composition strategy can integrate both local and global cues, and benefits the temporal action localization task.

However, the above attention composition strategy is still confused by the following challenges. First, given an untrimmed video, a large number of background frames would bring strong noises and disturb the estimation of the modality attention. Second, because the frame attention and the modality attention are independently inferred, simple attention composition cannot set constraints between these two weights and would lose the structure information. Third, proper regularizers should be proposed to assist the learning process. To alleviate above challenges, we propose the action-aware pooling strategy in subsection \ref{secActionAwarePooling} to precisely estimate the modality attention. Besides, subsection \ref{secOTRegularization} jointly justifies the modality attention and the frame attention via an optimal transport process, and constrains the attention weights via a learnable frame-modality structure matrix. Moreover, subsection \ref{secRegularizing} proposes two regularizers to assist the learning process. In the end, the complete training and inference processes are illustrated in subsection \ref{secTrainInfer}.

\subsection{Generating action-aware features}
\label{secActionAwarePooling}
Given video features $\mathbf{F} \in \mathbb{R}^{2D \times T}$, we can infer frame attention $\mathbf{a}^{\rm f}$ via a temporal convolutional layer. As for modality attention $\mathbf{a}^{\rm m}$, a natural way to obtain the holistic feature is global average pooling, \ie $\mathbf{f}^{\rm *}=\frac{1}{T} \sum^{T}_{t=1} \mathbf{F}[:,t]$. Then, the modality attention can be predicted via a convolutional layer.

In the above process, the global average pooling operation equally treats each frame's feature. However, to obtain precise modality attention, we should pay more attention to actions and suppress noises from backgrounds. Recently, Lee \etal \cite{lee2021weakly} have demonstrated that a feature vector with a larger magnitude is more likely to indicate an action instance. Thus, we propose an action-aware pooling strategy based on the magnitude of each feature vector. Specifically, we calculate the Euclidean norm of each feature vector, rank norms in descending order, and obtain the ranking indexes $\mathcal{X}_{v} \in \mathbb{R}^{1 \times T}$. Afterward, we use a ratio $k$ (\eg 0.125) and set the number of potential action features as $N_{k} = \lfloor T \cdot k \rfloor$, where $\lfloor \cdot \rfloor$ indicates the floor down operation. Next, we get indexes for potential action features as $\mathcal{X}=\mathcal{X}_{v}[1:N_{k}]$, and calculate the action-aware feature by average pooling among these $N_{k}$ features: 
\begin{equation}
\setlength{\abovedisplayskip}{3pt}
    \mathbf{f}^{\rm w}=\frac{1}{N_{k}} \sum^{N_{k}}_{t=1} \mathbf{f}_{\mathbb{I}_{t}},
\setlength{\belowdisplayskip}{3pt}
\end{equation}
where $\mathbb{I}_{t}$ is the index for the feature vector with the $t^{th}$ maximum Euclidean norm. Thus, we replace the global average pooling operation with this action-aware pooling operation to predict modality attention $\mathbf{a}^{\rm m}$.

\subsection{Justifying the attention assignment via optimal transport}
\label{secOTRegularization}

In this Section, we introduce an optimal transport process to justify the inferred modality attention and frame attention, as well as capturing the structured information. From the perspective of attention assignment, the modality attention $\mathbf{a}^{\rm m}$ can be regarded as the supplier and the frame attention $\mathbf{a}^{\rm f}$ can be regarded as the receiver. The assignment process aims to determine an optimal transport plan $\mathbf{\Psi}$ and transport attention from the modality supplier to the frame receiver. In the transport plan $\mathbf{\Psi}$, each element $\psi_{m,t}$ indicates that the $m^{th}$ modality transports $\psi_{m,t}$ attention weight to the $t^{th}$ frame.

\begin{algorithm*}[htb]
	
	\caption{Optimal Transport based Assignment Justification.}
	\label{algOTR}
	\begin{algorithmic}[1]
		\STATE \textbf{Input:} Structure matrix $\mathbf{S} \in \mathbb{R}^{2 \times T}$; Modality attention $\mathbf{a}^{\rm m} \in \mathbb{R}^{2 \times 1}$; Frame attention $\mathbf{a}^{\rm f} \in \mathbb{R}^{1 \times T}$; Interation number $\rm N$.
		\STATE \textbf{Output:} Optimal transport loss $\mathcal{L}_{\rm OT}$.
		\STATE Set $\mathbf{u}=\mathbf{1}\in \mathbb{R}^{2 \times 1}$, $\mathbf{v}=\mathbf{1} \in \mathbb{R}^{1 \times T}$, $\epsilon=1 \times 10^{-3}$, $\gamma=1 \times 10^{-8}$.
		\FOR{ $n$=1,2,...N}
		\STATE  $ \mathbf{v}=\mathbf{v} + \epsilon \cdot (\log{(\mathbf{a}^{\rm f} + \gamma}) - \log{\sum_{i=1}^{2}{\exp{((-\mathbf{S}+ \Phi(\mathbf{u}) + \Phi(\mathbf{v}))/\epsilon}})}) $ \ \ // $\Phi$: unsqueeze operation
		\STATE $\mathbf{u}= \mathbf{u} + \epsilon \cdot (\log{(\mathbf{a}^{\rm m} + \gamma}) - \log{\sum_{i=1}^{T}{\exp{((-\mathbf{S}+ \Phi(\mathbf{u}) +  \Phi(\mathbf{v}))/\epsilon}})}) $
		\ENDFOR
		
		\STATE $\mathbf{\Psi}^{*}=\exp{((-\mathbf{S}+ \Phi(\mathbf{u}) + \Phi(\mathbf{v}))/\epsilon})$  \ \ // $\mathbf{\Psi}^{*}$: optimal transport plan
		\STATE $\mathcal{L}_{\rm OT} = \sum\nolimits_{m = 1}^2 {\sum\nolimits_{t = 1}^T {{s_{m,t}} \cdot {\psi^{*} _{m,t}}}}$  \ \ // Cost to assign modality attention to each frame.
		\STATE Return $\mathcal{L}_{\rm OT}$
		
	\end{algorithmic}
\end{algorithm*}

In the assignment process, each attention transportation requires a specific cost. The cost information can capture the frame-modality structure and form the structure matrix $\mathbf{S} \in \mathbb{R}^{2 \times T}$, where an element $s_{m,t}$ indicates the cost to transport unit attention from the $m^{th}$ modality to the $t^{th}$ frame. Intuitively, it leads to little cost that transporting attention to a discriminative modality of an action frame, while it would cause high cost that transporting attention to an indiscriminative modality of a background frame. In the vanilla optimal transport research \cite{ge2021ota}, the structure matrix can be constructed by prior knowledge. However, this paradigm is not suitable for the studied frame-modality attention composition problem, because there is no prior knowledge about the structure matrix. To alleviate this issue, we estimate the structure matrix via a convolutional layer and capture the structure information in the optimization process.

Given structure matrix $\mathbf{S}$, modality attention $\mathbf{a}^{\rm m}$ and frame attention $\mathbf{a}^{\rm f}$, the attention assignment problem can be formulated as:
\begin{equation}
\setlength{\abovedisplayskip}{3pt}
\setlength{\belowdisplayskip}{3pt}
	\begin{split}
	\mathop {\min }\limits_\mathbf{\Psi}  &{\rm{  }}\sum\limits_{m = 1}^2 {\sum\limits_{t = 1}^T {{s_{m,t}} \cdot {\psi_{m,t}}} } \\
	{\rm{  }}s.t.{\rm{  }}&\sum\limits_{m = 1}^2 {{\psi_{m,t}} = {a}_t^{{\rm{f}}}} ,{\rm{  }}\sum\limits_{t = 1}^T {{\psi_{m,t}} = } {a}_m^{{\rm{m}}}, \ \ {\psi_{m,t}} \ge 0, \\
	&\ m \in [1,2], \ \ t \in [1,2,...,T],
	\end{split}
	\label{eqOT}
\end{equation}
where the constrain conditions indicate: (1) For the $t^{\rm th}$ frame, the sum of appearance attention and motion attention is equal to its frame attention. (2) For a modality, it assigns holistic attention to all $T$ frames. (3) The attention weight assigned from a modality to a frame is non-negative.

To solve Eq. (\ref{eqOT}), we follow existing works \cite{cuturi2013sinkhorn,chen2020graph, ge2021ota} and introduce an entropy-regularization term. Then, we can optimize the dual problem of Eq. (\ref{eqOT}) and achieve the optimal attention assignment via the Sinkhorn-Knopp algorithm \cite{cuturi2013sinkhorn}. Specifically, the entropy regularized optimal transport problem in Eq. (\ref{eqOT}) can be solved via calculating the dual-Sinkhorn divergence, which can be effectively integrated into the end-to-end learning process. In the implementation, we employ the \textit{logsumexp} strategy to stabilize the numerical calculation. To make this clear, we exhibit the algorithm to perform assignment justification via optimal transport in Algorithm \ref{algOTR}. When performing Sinkhorn-Knopp iteration, we set fixed iteration steps by following previous work \cite{ge2021ota}. Another feasible strategy is monitoring variations of $\mathbf{u}$ and $\mathbf{v}$ as in previous works \cite{cuturi2013sinkhorn, zhao2018label}, which stops iteration when variations are small enough. In our experiments, the fixed iteration strategy performs slightly superior.

By optimizing the optimal transportation loss, we can directly optimize the structure matrix $\mathbf{S}$ and implicitly optimize the modality attention $\mathbf{a}^{\rm m}$ and frame attention $\mathbf{a}^{\rm f}$. Then, the Sinkhorn-Knopp algorithm can find an optimal transport plan $\mathbf{\Psi}^{*}$ and calculate the transport loss $\mathcal{L}_{\rm OT}$. Based on $\mathcal{L}_{\rm OT}$, the optimal transport process captures structure information via the learnable structure matrix, and constrains the mutual relationship between the modality attention $\mathbf{a}^{\rm m}$ and the frame attention $\mathbf{a}^{\rm f}$.

\subsection{Regularizing the learning process}
\label{secRegularizing}

We introduce two regularizers to assist the learning of the optimal transport process.

\textbf{Smoothness regularizer.} As neighboring frames exhibit similar appearances and motion patterns, their assignment costs should be close. To this end, we first convert the structure matrix $\mathbf{S} \in \mathbb{R}^{2 \times T}$ to the structure vector $\mathbf{s}^{\rm f} \in \mathbb{R}^{1 \times T}$ via a max-pooling operation. Then, we calculate a neighboring difference vector $\mathbf{d}$ as $d_{t} = s^{\rm f}_{t+1} - s^{\rm f}_{t}$. Under temporal smoothness regularizer, most elements in the neighboring difference vector $\mathbf{d}$ should be small, while the difference elements around action boundaries should be significant. Consequently, we set a ratio $\eta$ and select minimal $N_{\eta} = \lfloor T \cdot \eta \rfloor$ elements. Then, we calculate the mean value $q^{\rm d}$ among these minimal $N_{\eta}$ elements:
\begin{equation}
\setlength{\abovedisplayskip}{3pt}
\setlength{\belowdisplayskip}{3pt}
q^{\rm d} = \frac{1}{N_{\eta}} \min _{\mathcal{Q} \subset [d_{1}, d_{2},...,d_{T-1}] \atop |\mathcal{Q}|=N_{\eta}} \sum_{t=1}^{N_{\eta}} \mathcal{Q}_{t}.
\end{equation}

Because the mean value $q^{\rm d}$ should be small, the smoothness loss can be calculated as:
\begin{equation}
\setlength{\abovedisplayskip}{3pt}
\setlength{\belowdisplayskip}{3pt}
\mathcal{L}_{\rm s} = - {\rm log} (1 - q^{\rm d}).
\end{equation}

\textbf{F-norm regularizer.} In the optimal transport process, to prevent the structure matrix from trivial solution (\eg all zeros), we apply the F-norm\cite{li2019discriminative} to the structure matrix and calculate the normalization loss $\mathcal{L}_{\rm F}$ as follows:
\begin{equation}
\setlength{\abovedisplayskip}{3pt}
\setlength{\belowdisplayskip}{3pt}
	\mathcal{L}_{\rm F}=- \frac{1}{2T} \sqrt{\sum_{m=1}^{2} \sum_{t=1}^{T} {s_{m,t}}^{2}}.
\end{equation}

\subsection{Structured attention composition}
\label{secTrainInfer}

Fig. \ref{figFramework} exhibits the complete framework of the proposed structured attention composition module. Given appearance features $\mathbf{F}^{\rm a}$ and motion features $\mathbf{F}^{\rm m}$, we can calculate video features as $\mathbf{F}^{\rm v}=[\Theta^{\rm a}(\mathbf{F}^{\rm a}), \Theta^{\rm m}(\mathbf{F}^{\rm m})]$, where $\Theta^{*}(\cdot)$ indicates transforming raw features via two convolutional layers, $[\cdot]$ is the concatenation operation. Then, the action-aware pooling process assists in predicting the modality attention, and the frame attention can be predicted via a convolutional layer.

Before assignment justification, we normalize the frame attention and the modality attention. In particular, given a video with temporal length $T$, we set the total quantity of attention as $T$. Based on this, the appearance and motion modality attention can be calculated as $T \cdot a^{\rm m}_{1}$ and $T \cdot a^{\rm m}_{2}$, respectively, where $a^{\rm m}_{1}$ and $a^{\rm m}_{2}$ indicate the first and the second element in the modality attention $\mathbf{a}^{\rm m}$. The frame attention $\mathbf{a}^{\rm f}$ can be calculated as $a^{\rm f}_{t} = T \cdot \frac{a^{\rm f}_{t}}{\sum_{t=1}^{T} a^{\rm f}_{t}}$.

The proposed structured attention composition module is a plug-and-play module for off-the-shelf temporal action localization methods. Specifically, given vanilla appearance features and motion features, it predicts frame-modality attention and modulates the vanilla features, which can be fed into the off-the-shelf action localization methods. During training, in addition to existing loss functions of each method, the structured attention composition module calculates an extra loss including three terms: the optimal transport loss $\mathcal{L}_{\rm OT}$, the temporal smoothness loss $\mathcal{L}_{\rm s}$ and the F-norm loss $\mathcal{L}_{\rm F}$:
\begin{equation}
\setlength{\abovedisplayskip}{3pt}
\setlength{\belowdisplayskip}{3pt}
    \mathcal{L} = \mathcal{L}_{\rm OT} + \lambda_{\rm s} \cdot \mathcal{L}_{\rm s} + \lambda_{\rm F} \cdot \mathcal{L}_{\rm F},
    \label{edLoss}
\end{equation}
where $\lambda_{\rm s}$ and $\lambda_{\rm F}$ are trade-off coefficients. In inference, we first predict frame attention and modality attention, then compose the frame-modality attention. Afterward, we employ the frame-modality attention to modulate appearance and motion features, and localize action instances via off-the-shelf methods.

\section{Experiments}

\subsection{Experimental setups}
\label{secExpSetups}

\textbf{Dataset.} We perform experiments on two widely used benchmark datasets THUMOS14 \cite{THUMOS14} and ActivityNet v1.3 \cite{caba2015activitynet}. THUMOS14 \cite{THUMOS14} consists of 200 validation videos and 213 testing videos, belonging to 20 action categories. Challenges on THUMOS14 lies in motion blur, drastic varieties of action duration (from fractions of a second to tens of seconds), dense action instances (each validation video contains 15 instances in average), \etc We keep identical setups with previous works \cite{liu2021multi, lin2021learning, xu2020g}, learn the proposed structured attention composition module on the validation set and evaluate the performance on the testing set. ActivityNet v1.3 \cite{caba2015activitynet} contains 200 categories and 19994 videos, where the ratio of training, validation and testing videos is 2:1:1. Challenges on ActivityNet v1.3 lie in numerous action categories, extremely long action instances (\eg hundreds of seconds), drastic intra-class variations caused by camera shift. We follow previous works \cite{yaothree, lin2021learning, liu2021multi} and employ the I3D model \cite{carreira2017quo} pre-trained on Kinetics-400 dataset to extract appearance and motion features. Specifically, we employ off-the-shelf I3D model and do not conduct finetuning on benchmark datasets \cite{THUMOS14, caba2015activitynet}, where the optical flow is calculated by the TV-L1 \cite{zach2007duality} algorithm.

\textbf{Metric.} We follow existing works \cite{wang2016temporal, liu2021multi} and employ the standard evaluation metric, interpolated average precision, to measure the performance of each method. Considering the $c^{th}$ action category with $N_{c}$ predicted action instances, we rank all predictions according to the descending order of the confidence score, and calculate precision $p_{i}$ for each instance. Then, the average precision $AP_{c}$ of the $c^{th}$ category can be calculated as:
\begin{equation}
  AP_{c}=\frac{1}{N_{c}}\sum^{N_{c}}_{n=1} \mathop{\rm max}\limits_{i \geq n}\ p_{i}.
\end{equation}
Next, the mean value of average precision across all categories gives the complete score:
\begin{equation}
	mAP=\frac{1}{C} \sum_{c=1}^{C} AP_{c}.
\end{equation}
As for THUMOS14, we report mAP under seven IoU thresholds (0.1:0.7) for comprehensive analysis, where the official evaluation metric focuses on mAP under IoU threshold 0.5. As for ActivityNet v1.3, the evaluation metric is the average mAP under ten evenly distributed IoU thresholds from 0.50 to 0.95.

Considering the randomicity in the initialization and optimization processes, the performance of an action localization algorithm may vary a bit among different experiments. To alleviate this issue and verify the robust gains brought by the proposed structured attention composition module, we carry out each experiment 6 times and report the mean value. Moreover, we utilize the Student's $t$-test to check the differences between two experimental setups: the existing method, the existing method equipped with the structured attention composition module:
\begin{equation}
    t=\frac{\bar{S}_{1}-\bar{S}_{2}}{\sigma_{p} \cdot \sqrt{\frac{1}{n_{1}}+\frac{1}{n_{2}}}},
    \label{eqTTest}
\end{equation}
where $\bar{S}_{1}$ and $\bar{S}_{2}$ indicate mean values, $n_{1}$ and $n_{2}$ are experiment numbers, and $\sigma_{p}$ represents the pooled standard deviation of two groups of results, which can be calculated as:
\begin{equation}
    \sigma_{p}=\sqrt{\frac{\left(n_{1}-1\right) \sigma_{S_{1}}^{2}+\left(n_{2}-1\right) \sigma_{S_{2}}^{2}}{n_{1}+n_{2}-2}},
    \label{eqSigma}
\end{equation}
where $\sigma_{S_{1}}^{2}$ and $\sigma_{S_{2}}^{2}$ indicate the variance of scores. In theory, if $p$-value is smaller than 0.05, it is confident to conclude that the proposed structured attention composition module can bring noticeable performance gains to the existing action localization methods.

\begin{table}[thbp]
	\centering
	\caption{Efficacy of the proposed structured attention composition (SAC) module for the temporal action localization task, measured by mAP (\%) on THUMOS14 dataset. Avg. indicates the average mAP (\%) among thresholds 0.1:0.7.}
	\setlength{\tabcolsep}{5pt}
	\begin{tabular}{c|ccccccc|c}
    \toprule
    \multirow{2}[4]{*}{Mehtod} & \multicolumn{7}{c|}{mAP@IoU (\%)}                     & Avg. \\
\cmidrule{2-8}          & 0.1   & 0.2   & 0.3   & 0.4   & 0.5   & 0.6   & 0.7   & (\%) \\
    \midrule
    G-TAD \cite{xu2020g} & 65.0  & 62.5  & 58.0  & 51.8  & 42.9  & 33.2  & 23.1  & 48.1 \\
    G-TAD+SAC & 65.5  & 62.3  & 59.0  & 52.4  & 44.4  & 34.8  & 24.2  & 48.9 \\
    Improvement & \textbf{+0.5} & -0.2  & \textbf{+1.0} & \textbf{+0.6} & \textbf{+1.5} & \textbf{+1.6} & \textbf{+1.1} & \textbf{+0.8} \\
    \midrule
    SSAD \cite{lin2017single} & 72.2  & 69.7  & 65.2  & 56.5  & 45.5  & 31.0  & 17.0  & 51.0 \\
    SSAD+SAC & 75.5  & 73.7  & 69.1  & 61.1  & 51.4  & 37.1  & 22.8  & 55.8 \\
    Improvement & \textbf{+3.3} & \textbf{+4.0} & \textbf{+3.9} & \textbf{4.6} & \textbf{+5.9} & \textbf{+6.1} & \textbf{+5.8} & \textbf{+4.8} \\
    \midrule
    \cite{liu2021multi} & 73.3  & 71.9  & 68.7  & 64.1  & 56.8  & 45.9  & 30.1  & 58.7 \\
    \cite{liu2021multi}+SAC & 74.9  & 73.2  & 69.3  & 64.8  & 57.6  & 47.0  & 31.5  & 59.8 \\
    Improvement & \textbf{+1.1} & \textbf{+1.0} & \textbf{+0.4} & \textbf{+0.8} & \textbf{+0.7} & \textbf{+0.7} & \textbf{+0.5} & \textbf{+1.1} \\
    \bottomrule
    \end{tabular}%
	\label{tabImpTalThumos}%
\end{table}%

\textbf{Implementation details.} We carry out experiments on four well-performed methods SSAD \cite{lin2017single}, G-TAD \cite{xu2020g}, BMN \cite{lin2019bmn} and Liu \etal \cite{liu2021multi}, to verify the efficacy of the proposed structured attention composition module. Among these four methods, we use publicly available official implementations for G-TAD \cite{xu2020g}\footnote{\url{https://github.com/frostinassiky/gtad}} and \cite{liu2021multi}\footnote{\url{https://github.com/xlliu7/MUSES}}, utilize the implementation of JJBOY\footnote{\url{https://github.com/JJBOY/BMN-Boundary-Matching-Network}} for BMN \cite{lin2019bmn}. As SSAD\cite{lin2017single} lacks the official implementation, we re-implement it and employ the I3D network\cite{carreira2017quo} pre-trained on the Kinetics-400 dataset to extract features, following recent works\cite{liu2021multi, lin2021learning, zeng2021graph}.  As for G-TAD \cite{xu2020g}, BMN \cite{lin2019bmn} and \cite{liu2021multi}, we try to keep the original experimental setups unchanged to pursue fair comparisons, \eg employing the provided features and using the same hyper-parameters. In the implementation of the structured attention composition module, we set the ratio $\eta=80\%$ to select elements and set $k=1/8$ in the action-aware pooling process. The proposed structured attention composition module is introduced into the above four methods \cite{lin2017single, xu2020g, lin2019bmn, liu2021multi}, and consistently improve action localization performance. All experiments are performed on a workstation equipped with 2080 Ti GPUs.

\begin{figure*}
     \centering
     \begin{subfigure}[b]{0.315\textwidth}
         \centering
         \graphicspath{{figure/}}
         \includegraphics[width=\textwidth]{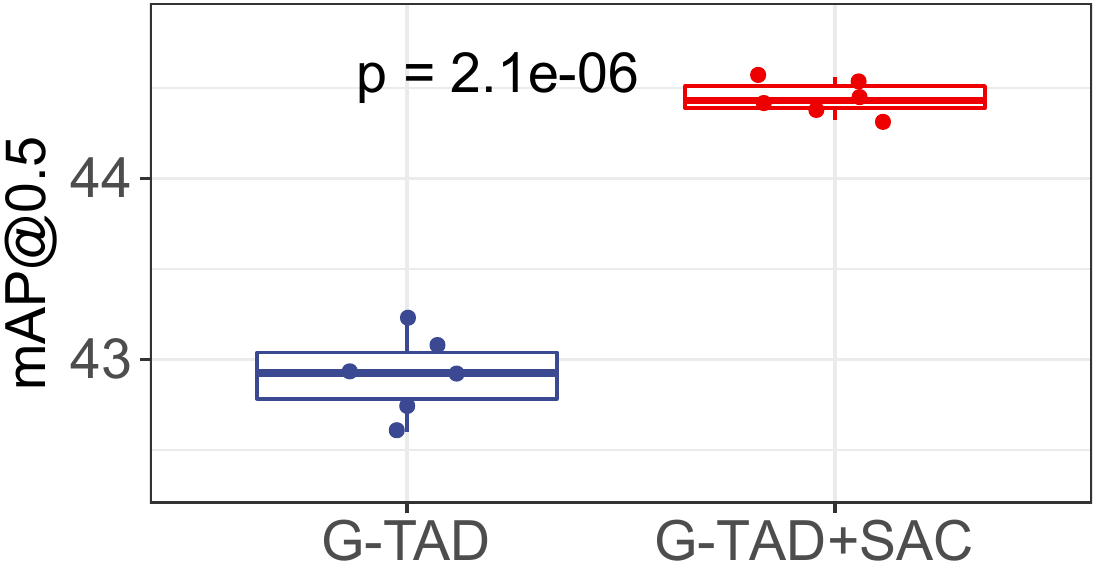}
         \caption{G-TAD \cite{xu2020g} on THUMOS14}
         \label{figThumosG-TAD}
     \end{subfigure}
     \begin{subfigure}[b]{0.315\textwidth}
         \centering
         \graphicspath{{figure/}}
         \includegraphics[width=\textwidth]{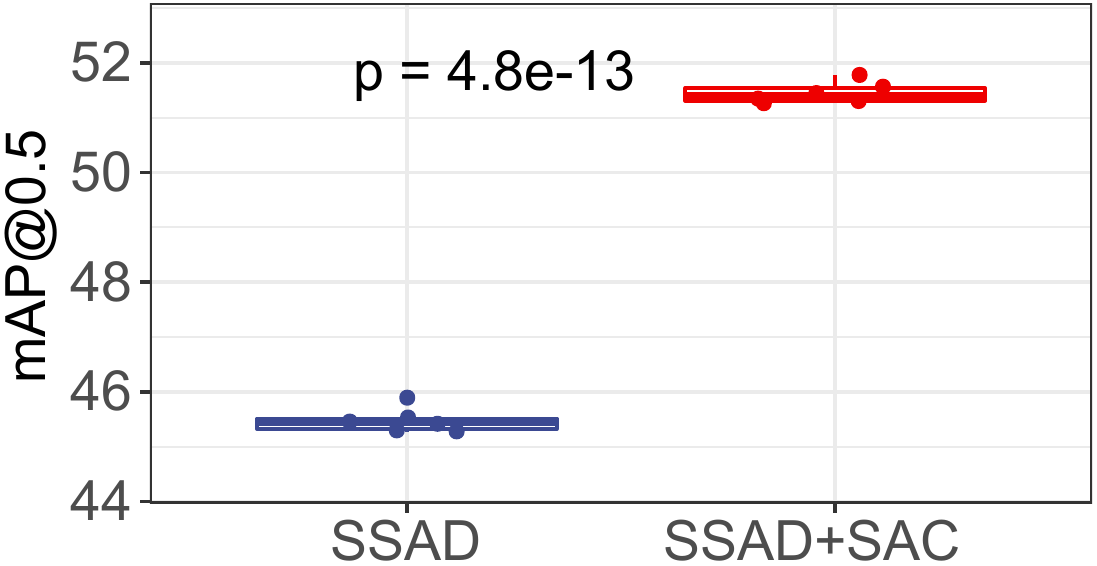}
         \caption{SSAD \cite{lin2017single} on THUMOS14}
         \label{figThumosSSAD}
     \end{subfigure}
     \begin{subfigure}[b]{0.315\textwidth}
         \centering
         \graphicspath{{figure/}}
         \includegraphics[width=\textwidth]{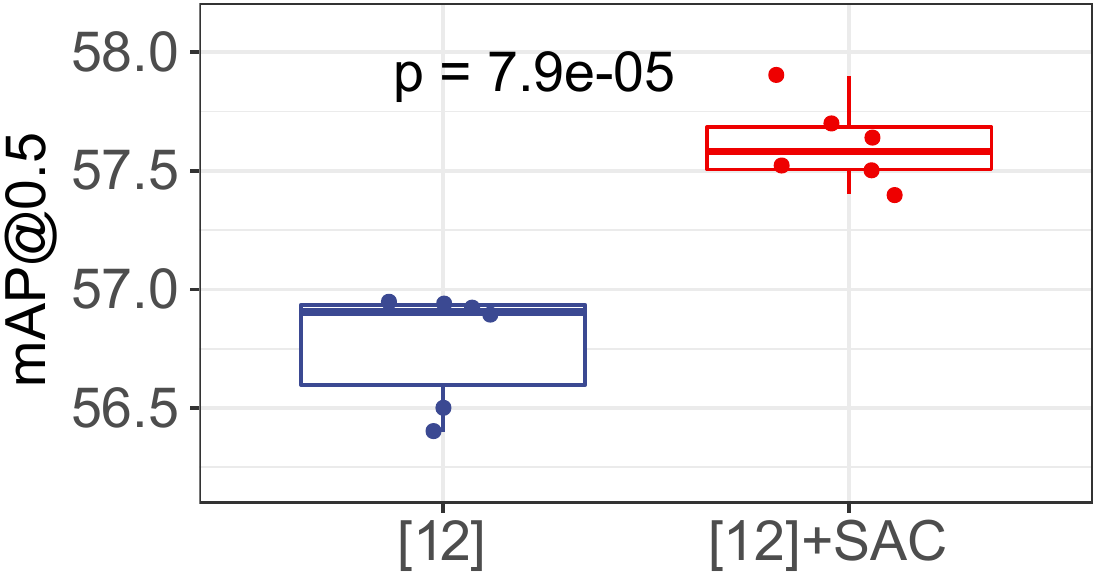}
         \caption{\cite{liu2021multi} on THUMOS14}
         \label{figThumosMUSES}
     \end{subfigure}
     \begin{subfigure}[b]{0.315\textwidth}
         \centering
         \graphicspath{{figure/}}
         \includegraphics[width=\textwidth]{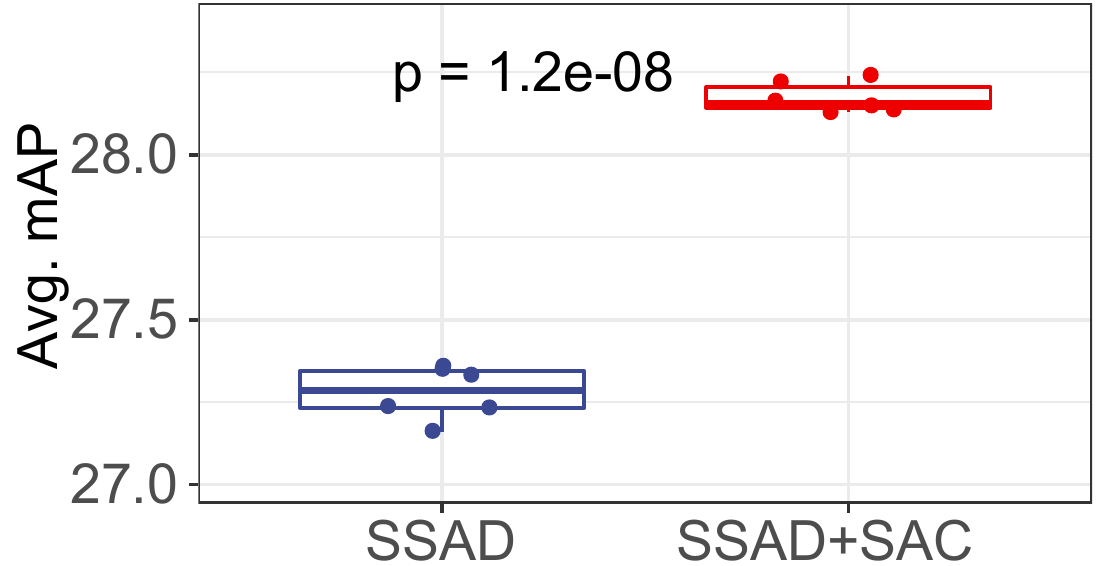}
         \caption{SSAD \cite{lin2017single} on ActivityNet v1.3}
         \label{figANetSSAD}
     \end{subfigure}
     \begin{subfigure}[b]{0.315\textwidth}
         \centering
         \graphicspath{{figure/}}
         \includegraphics[width=\textwidth]{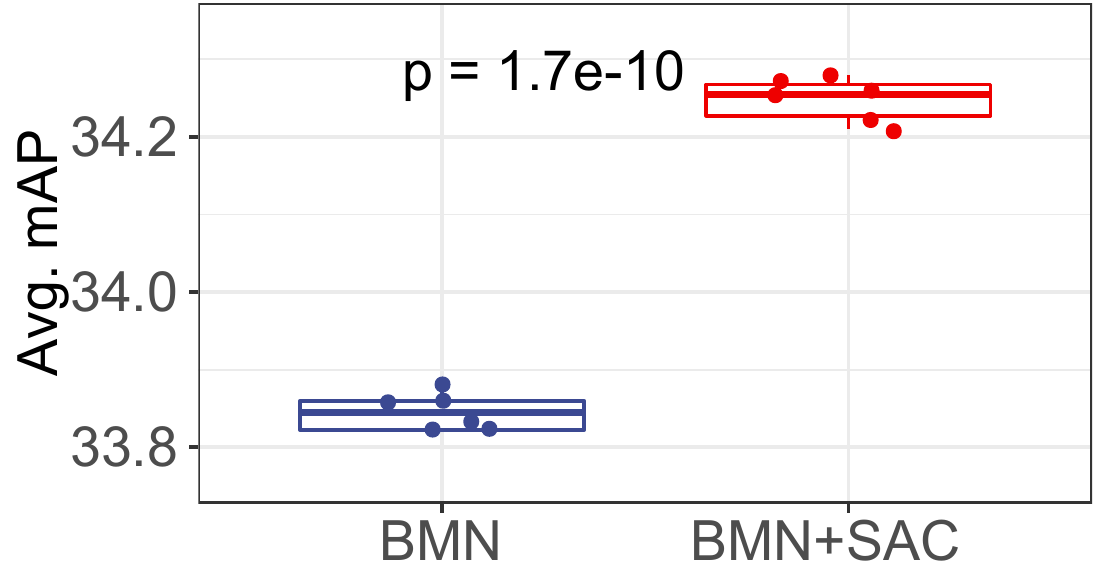}
         \caption{BMN \cite{lin2019bmn} on ActivityNet v1.3}
         \label{figANetBMN}
     \end{subfigure}
     \begin{subfigure}[b]{0.315\textwidth}
         \centering
         \graphicspath{{figure/}}
         \includegraphics[width=\textwidth]{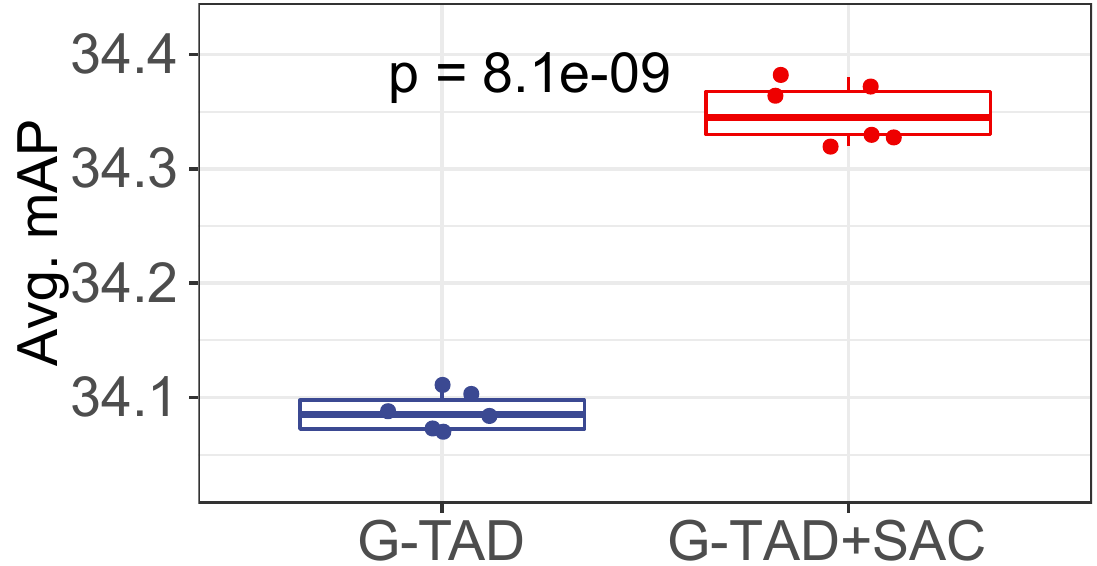}
         \caption{G-TAD \cite{xu2020g} on ActivityNet v1.3}
         \label{figANetGTAD}
     \end{subfigure}
        \caption{Detailed performance analysis between vanilla methods \cite{xu2020g, lin2017single, liu2021multi, lin2019bmn} (shown in blue) and the ones assisted with our proposed structured attention composition module (shown in red). As t-test gives small $p$-values, it is confident to conclude that the structured attention composition module can consistently improve the performance of these methods.}
        \label{figPerfBar}
        \vspace{-0.3cm}
\end{figure*}

\subsection{Improvements over different baselines}
The proposed structured attention composition module predicts frame-modality attention to highlight discriminative modality of action features. As a plug-and-play module, it can assist existing temporal action localization methods and improve localization accuracy. We start from a brief review of representative temporal action localization methods \cite{lin2017single, lin2019bmn, xu2020g, liu2021multi}:
\begin{itemize}
    \item SSAD \cite{lin2017single} employs the anchor mechanism, utilizes a pyramid network to capture action instances with various durations, and performs one-stage action localization.
    \item BMN \cite{lin2019bmn} performs frame-level classification to obtain action proposals, and estimates the confidence score of each proposal via the boundary matching mechanism.
    \item G-TAD \cite{xu2020g} emphasizes the importance of semantic context, and utilizes the graph convolution to learn from temporal neighbors and aggregate semantic contexts.
    \item Liu \etal \cite{liu2021multi} propose a temporal aggregation module to enhance the feature coherence within an action instance, then perform detection via classifying proposals.
\end{itemize}
In above four methods, some works \cite{lin2017single, lin2019bmn, xu2020g} adopt the early-fusion strategy that concatenates appearance features and motion features to represent the input video, while the other work \cite{liu2021multi} adopts the late-fusion strategy that balances the influence of appearance features and motion features via a fixed coefficient. To explicitly distinguish the influence of each frame and each modality, we introduce the structured attention composition module to these works, and precisely learn frame-modality attention. We select three well-performed methods \cite{xu2020g, lin2017single, liu2021multi} to carry out experiments on THUMOS14, while \cite{lin2017single, lin2019bmn, xu2020g} are used for experiments on ActivityNet v1.3 to verify the effectiveness of the proposed structured attention composition module.

Table \ref{tabImpTalThumos} reports improvements brought by the structured attention composition module on the THUMOS14 dataset. Considering the official metric, mAP under threshold 0.5, the structured attention composition module consistently improves the performance of three well-performed methods \cite{xu2020g, lin2017single, liu2021multi}, and brings 1.5\%, 5.9\% and 0.7\% mAP improvements for \cite{xu2020g}, \cite{lin2017single} and \cite{liu2021multi}, respectively. Considering the average mAP under thresholds 0.1:0.7, the structured attention composition module brings 0.8\%, 4.8\% and 1.1\% mAP improvements as well. As a classical temporal action localization method, SSAD \cite{lin2017single} directly utilizes the anchor mechanism to localize action instances, without considering frame and modality attention. The apparent improvements brought by our structured attention composition module reveals that explicitly modeling frame-modality attention is crucial to temporal action localization. Although improvements of \cite{xu2020g} and \cite{liu2021multi} are not as obvious as the one of SSAD \cite{lin2017single}, it is worth noting that we keep all hyper-parameters unchanged and consistently observe performance gains under multiple IoU thresholds. It is possible to further improve the performance after careful parameter tuning, which can be studied in future works.

\begin{table}[thbp]
	\centering
	\caption{Efficacy of the proposed structured attention composition module for the temporal action localization task, measured by mAP (\%) on ActivityNet v1.3 dataset.}
	\setlength{\tabcolsep}{7pt}
	\begin{tabular}{c|ccc|c}
    \toprule
    \multirow{2}[4]{*}{Mehtod} & \multicolumn{3}{c|}{mAP@IoU (\%)} & \multirow{2}[4]{*}{Average mAP (\%)} \\
\cmidrule{2-4}          & 0.50  & 0.75  & 0.95  &  \\
    \midrule
    SSAD \cite{lin2017single} & 43.85 & 27.84 & 5.17  & 27.28 \\
    SSAD \cite{lin2017single}+SAC & 44.88 & 28.96 & 5.28  & 28.17 \\
    Improvement & \textbf{+1.03} & \textbf{+1.12} & \textbf{+0.11} & \textbf{+0.89} \\
    \midrule
    BMN \cite{lin2019bmn} & 50.07 & 34.78 & 8.29  & 33.85 \\
    BMN \cite{lin2019bmn}+SAC & 50.42 & 35.26 & 7.92  & 34.25 \\
    Improvement & \textbf{+0.35} & \textbf{+0.48} & -0.37 & \textbf{+0.40} \\
    \midrule
    G-TAD \cite{xu2020g} & 50.36 & 34.60 & 9.02  & 34.09 \\
    G-TAD \cite{xu2020g}+SAC & 50.73 & 35.43 & 7.29  & 34.35 \\
    Improvement & \textbf{+0.37} & \textbf{+0.83} & -1.73 & \textbf{+0.26} \\
    \bottomrule
    \end{tabular}%
    \vspace{-0.3cm}
	\label{tabImpTalAnet}%
\end{table}

Table \ref{tabImpTalAnet} reports the efficacy of the structured attention composition module on the ActivityNet v1.3 dataset. Under the metric average mAP, SSAD \cite{lin2017single}, BMN \cite{lin2019bmn} and G-TAD \cite{xu2020g} receive 0.89\%, 0.40\% and 0.26\% mAP improvements, respectively. Because the evaluation on ActivityNet v1.3 adopts strict IoU thresholds (\ie 0.50:0.95), these improvements are sufficient to demonstrate the effectiveness of the proposed structured attention composition module. In addition, we observe that the number of predicted action instances would decrease when a method is equipped with the structured attention composition module. To our best knowledge, the algorithm with decreased prediction number is prone to lose some action instances, which is the potential reason that the performance gains under IoU threshold 0.95 are small or even negative. However, considering the metric average mAP, the structured attention module benefits action localization on the ActivityNet v1.3 dataset.

To verify the robust gains brought by the structured attention composition module, we carry out each experiment 6 times and report the detailed performance in Fig. \ref{figPerfBar}. In each boxplot, the Student's $t$-test gives small $p$-values, \eg based on SSAD \cite{lin2017single}, the $p$-value is as small as $4.8 \times 10^{-13}$. These small $p$-values demonstrate that the proposed structured attention composition module can steadily improve the performance of these action localization methods.

\begin{table}[thbp]
  \centering
  \caption{Comparison experiments on THUMOS14 dataset, measured by mAP (\%) under different IoU thresholds. Our method introduces the structured attention composition module into \cite{liu2021multi}. Avg. indicates the average mAP (\%) among thresholds 0.3:0.7.}
  \setlength{\tabcolsep}{5.5pt}
    \begin{tabular}{c|ccccc|c}
    \toprule
    \multirow{2}[4]{*}{Method} & \multicolumn{5}{c|}{mAP(\%)@IoU}      & Avg. \\
\cmidrule{2-6}          & 0.3   & 0.4   & 0.5   & 0.6   & 0.7   & (\%) \\
    \midrule
    GTAN\cite{long2019gaussian}{\tiny CVPR19} & 57.8  & 47.2  & 38.8  & -     & -     & - \\
    BMN\cite{lin2019bmn}{\tiny ICCV19} & 56.0  & 47.4  & 38.8  & 29.7  & 20.5  & 38.5  \\
    BSN++\cite{su2020bsnPP}{\tiny AAAI21} & 59.9  & 49.5  & 41.3  & 31.9  & 22.8  & 41.1  \\
    TCANet\cite{qing2021temporal}{\tiny CVPR21} & 60.6  & 53.2  & 44.6  & 36.8  & 26.7  & 44.4  \\
    BUMR\cite{zhao2020bottom}{\tiny ECCV20} & 53.9  & 50.7  & 45.4  & 38.0  & 28.5  & 43.3  \\
    A2Net\cite{2020Revisiting}{\tiny TIP20} & 58.6  & 54.1  & 45.5  & 32.5  & 17.2  & 41.6  \\
    PGCN\cite{zeng2019graph}{\tiny ICCV19} & 63.6  & 57.8  & 49.1  & -     & -     & - \\
    PCG-TAL\cite{su2020pcg}{\tiny TIP21} & 65.1  & 59.5  & 51.2  & -     & -     & - \\
    PBRNet\cite{liu2020progressive}{\tiny AAAI20} & 58.6  & 54.6  & 51.3  & 41.8  & 29.5  & 47.2  \\
    G-TAD\cite{xu2020g}{\tiny CVPR20} & 66.4  & 60.4  & 51.6  & 37.6  & 22.9  & 47.8  \\
    RTD-Net\cite{tan2021relaxed}{\tiny ICCV21} & 68.3  & 62.3  & 51.9  & 38.8  & 23.7  & 49.0  \\
    GCM\cite{zeng2021graph}{\tiny TPAMI21} & 66.5  & 60.8  & 51.9  & -     & -     & - \\
    C-TCN\cite{li2020deep}{\tiny ACMMM20} & 68.0  & 62.3  & 52.1  & -     & -     & - \\
    ContextLoc\cite{zhu2021enriching}{\tiny ICCV21} & 68.3  & 63.8  & 54.3  & 41.8  & 26.2  & 50.9  \\
    Li \etal\cite{yaothree}{\tiny CVPR21} & 63.2  & 58.5  & 54.8  & 44.3  & \textbf{32.4} & 50.6  \\
    AFSD\cite{lin2021learning}{\tiny CVPR21} & 67.3  & 62.4  & 55.5  & 43.7  & 31.1  & 52.0  \\
    Liu \etal\cite{liu2021multi}{\tiny CVPR21} & 68.9  & 64.0  & 56.9  & 46.3  & 31.0  & 53.4  \\
    \midrule
    Ours  & \textbf{69.3} & \textbf{64.8} & \textbf{57.6} & \textbf{47.0} & 31.5  & \textbf{54.0} \\
    \bottomrule
    \end{tabular}%
  \label{tabCmpTHUMOS}%
  \vspace{-0.5cm}
\end{table}%

\subsection{Comparison experiments}

We introduce the structured attention composition module to recent methods \cite{liu2021multi,xu2020g} and make comparisons with current state-of-the-art methods. As shown in Table \ref{tabCmpTHUMOS}, on the THUMOS14 benchmark, the structured attention composition module equipped with \cite{liu2021multi} builds new state-of-the-art performance, \ie an mAP of 57.6\% under IoU threshold 0.5. Besides, the average mAPs under thresholds 0.3:0.7 also verify the superiority of our method. In particular, compared with the most recent works \cite{tan2021relaxed, zhu2021enriching}, our method exceeds ContextLoc \cite{zhu2021enriching} with an mAP of 3.3\%; compared with works \cite{2020Revisiting, su2020pcg, zeng2021graph} published in prestigious journals, our method exceeds GCM \cite{zeng2021graph} with an mAP of 5.7\%. Considering the most competitive method \cite{liu2021multi}, where Liu \etal elaborately model both short-term and long-term relationships to enhance the feature coherence, the proposed structured attention composition module can bring an mAP of 0.7\% improvement, demonstrating the necessity of modeling the frame-modality attention. It can be noticed that \cite{yaothree} performs well under IoU threshold 0.7. Actually, \cite{yaothree} designs two pretext tasks (\ie multi-action classification and localization confidence estimation) to enhance the feature representation and improve the localization results. Although pretext tasks can benefit localization results, they would complicate training.

\begin{table}[htbp]
  \centering
  \caption{Comparison experiments on ActivityNet v1.3 datasets, measured by mAP (\%) under different IoU thresholds. Our method introduces the structured attention composition module into \cite{xu2020g}.}
    \begin{tabular}{c|cccc}
    \toprule
    \multirow{2}[4]{*}{Method} & \multicolumn{3}{c}{ActivityNet v1.3} & \multirow{2}[4]{*}{Average mAP (\%)} \\
\cmidrule{2-4}          & 0.50  & 0.75  & 0.95  &  \\
    \midrule
    \multicolumn{5}{c}{Methods using extra video-level labels} \\
    \midrule
    BMN\cite{lin2019bmn}{\tiny ICCV19} & 50.07 & 34.78 & 8.29  & 33.85 \\
    PGCN\cite{zeng2019graph}{\tiny ICCV19} & 48.26 & 33.16 & 3.27  & 31.11 \\
    BUMR\cite{zhao2020bottom}{\tiny ECCV20} & 43.47 & 33.91 & 9.21  & 30.12 \\
    RTD-Net\cite{tan2021relaxed}{\tiny ICCV21} & 46.43 & 30.45 & 8.64  & 30.46 \\
    ContextLoc\cite{zhu2021enriching}{\tiny ICCV21} & 56.01 & 35.19 & 3.55  & 34.23 \\
    AFSD\cite{lin2021learning}{\tiny CVPR21} & 52.40 & 35.30 & 6.50  & 34.40 \\
    BSN++\cite{su2020bsnPP}{\tiny AAAI21} & 51.27 & 35.70 & 8.33  & 34.88 \\
    PCG-TAL\cite{su2020pcg}{\tiny TIP21} & 52.04 & 35.92 & 7.97  & 34.91 \\
    PBRNet\cite{liu2020progressive}{\tiny AAAI20} & 53.96 & 34.97 & 8.98  & 35.01 \\
    \midrule
    \multicolumn{5}{c}{Methods using extra action proposals} \\
    \midrule
    Liu \etal\cite{liu2021multi}{\tiny CVPR21} & 50.02 & 34.97 & 6.57  & 33.99 \\
    GCM\cite{zeng2021graph}{\tiny TPAMI21} & 51.03 & 35.17 & 7.44  & 34.24 \\
    Li \etal\cite{yaothree}{\tiny CVPR21} & \textbf{57.80} & \textbf{37.60} & \textbf{9.60} & 35.00 \\
    TCANet\cite{qing2021temporal}{\tiny CVPR21} & 52.27 & 36.73 & 6.86  & \textbf{35.52} \\
    \midrule
    \multicolumn{5}{c}{Methods directly localizing actions} \\
    \midrule
    A2Net\cite{2020Revisiting}{\tiny TIP20} & 43.55 & 28.69 & 3.70  & 27.75 \\
    C-TCN\cite{li2020deep}{\tiny ACMMM20} & 47.60 & 31.90 & 6.20  & 31.10 \\
    G-TAD\cite{xu2020g}{\tiny CVPR20} & 50.36 & 34.60 & 9.02  & 34.09 \\
    GTAN\cite{long2019gaussian}{\tiny CVPR19} & 52.61 & 34.14 & 8.91  & 34.31 \\
    \midrule
    Ours   & 50.73 & 35.43 & 7.29  & 34.35 \\
    \bottomrule
    \end{tabular}%
  \label{tabCmpANet}%
\end{table}%

Table \ref{tabCmpANet} reports the comparison experiments on the ActivityNet v1.3 dataset. It can be seen that the structured attention composition module equipped with \cite{xu2020g} achieves comparable performance with recent methods under the metric average mAP. Specifically, some works use extra video-level classification labels to localize action instances. For example, PBRNet\cite{liu2020progressive} and BSN++\cite{su2020bsnPP} rely on CUHK \cite{zhao2017cuhk}, while PCG-TAL\cite{su2020pcg} and AFSD\cite{lin2021learning} utilize UntrimmedNet \cite{wang2017untrimmednets}. Besides, some works use extra action proposals. For example, TCANet\cite{qing2021temporal} relies proposals from BMN \cite{lin2019bmn}, while Li \etal \cite{yaothree} utilizes PBRNet\cite{liu2020progressive} to generate action proposals. In contrast to above works, the proposed method can directly discover action instances from untrimmed videos, and achieve comparable performance.

\subsection{Ablation experiments}
\label{secAblExp}

\begin{figure*}[htbp]
	\graphicspath{{figure/}}
	\centering
	\includegraphics[width=1.0\linewidth]{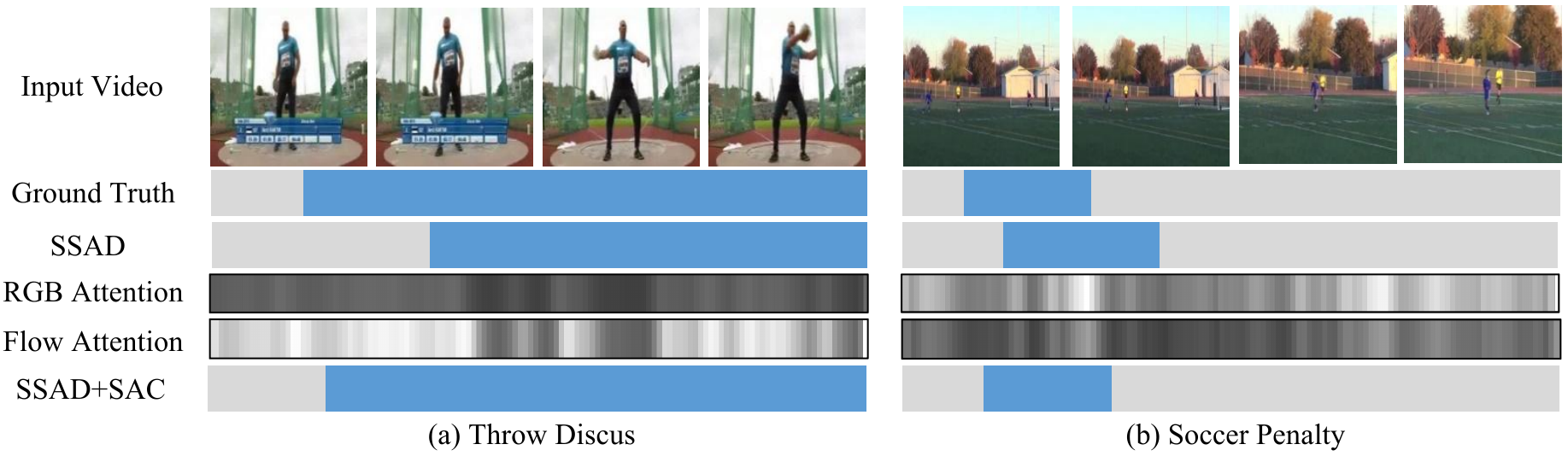}
	\caption{Qualitative results of the structured attention composition module, where lighter color indicates larger weight.}
	\label{figVisualization}
	\vspace{-0.5cm}
\end{figure*}

\begin{table}[thbp]
    \centering
    \setlength{\tabcolsep}{16pt}
    \caption{Ablation studies about different ways to infer frame-modality attention, measured by mAP (\%) on THUMOS14 dataset.}
    \begin{tabular}{l|cc}
    \toprule
    Setups & SSAD\cite{lin2017single} & G-TAD\cite{xu2020g} \\
    \midrule
    Baseline & 45.5  & 42.9 \\
    + Predicted Attention & 48.1  & 43.4 \\
    + Modality Attention & 47.5  & 43.2 \\
    + Frame Attention & 47.8  & 43.3 \\
    + Composed Attention & 48.8  & 43.7 \\
    \bottomrule
    \end{tabular}%
    \label{tabAblAtt}
    \vspace{-0.3cm}
\end{table}

\textbf{Effectiveness of the composed attention.} 
Table \ref{tabAblAtt} studies different ways to estimate the frame-modality attention. Under IoU threshold 0.5, the baseline method SSAD \cite{lin2017single} gets an mAP of 45.5\%. If we jointly consider the appearance feature and the motion feature (\eg via concatenation), and predict the frame-modality attention based on a convolutional layer, the performance can be improved to 48.1\% mAP. For one thing, it reveals the necessity to distinguish the influence of each frame and each modality; for another thing, the localization results are not accurate enough due to lack of modeling frame-modality relationship. Only predicting modality attention or frame attention achieve 47.5\% and 47.8\%, respectively. Compared with the performance of SSAD \cite{lin2017single}, this demonstrates that both the modality attention and the frame attention are influential to localizing action instances. Furthermore, a simple implementation of the composed attention strategy achieves an mAP of 48.8\%, demonstrating the promising aspects of learning frame-modality attention.

\begin{table}[thbp]
    \centering
    \setlength{\tabcolsep}{13pt}
    \caption{Efficacy of each component in the proposed structured attention composition module, measured by mAP (\%) on THUMOS14 dataset.}
    \begin{tabular}{l|cc}
    \toprule
    Setups & SSAD\cite{lin2017single} & G-TAD\cite{xu2020g} \\
    \midrule
    Baseline + Composed Attention & 48.8  & 43.7 \\
    + OT  Regularization & 50.0  & 44.0 \\
    + OT + Action-Aware Pooling & 50.9  & 44.3 \\
    + OT + Smooth Term & 50.8  & 44.2 \\
    Complete Method & 51.4  & 44.4 \\
    \bottomrule
    \end{tabular}%
    \label{tabAblSAC}
    \vspace{-0.3cm}
\end{table}

\textbf{Effectiveness of each module.} Table \ref{tabAblSAC} reports the performance gains brought by the optimal transport process, the action-aware pooling module and the smoothness regularizer. Compared with the vanilla composed attention, the optimal transport process measures the fitness between the frame attention and the modality attention. Specifically, the learnable structure matrix can estimate the transport cost from the modality attention to the frame attention, and guides the algorithm to obtain precise frame-modality attention weights. Based on SSAD \cite{lin2017single}, the optimal transport process brings an mAP of 1.2\% performance gains under IoU threshold 0.5. Based on this, the action-aware pooling strategy focuses on action features to predict the modality attention, which brings an mAP of 0.9\% gain. Besides, the smoothness regularizer guides the frame-modality structure matrix to be smooth among neighboring frames, which carries 0.8\% mAP gain. Our complete method achieves an mAP of 51.4\%, and the proposed structured attention composition module brings an mAP of 5.9\% performance gains over the vanilla SSAD \cite{lin2017single} method (\ie 45.5\%).

\textbf{Superiority over traditional attention mechanisms.} The core of this work is to learn precise frame-modality attention. Alternatively, two traditional attention mechanisms need to be compared:
\begin{itemize}
    \item Sequential attention. CBAM \cite{woo2018cbam} sequentially calculates the channel attention and the spatial attention to exploit the inter-channel and inter-spatial relationships. Similarly, we can first estimate the modality attention and then learn the frame attention, to model the frame-modality relationship.\footnote{We also conduct experiments that first estimate the frame attention, and then learn the modality attention, but observe a bit of performance drop.}
\end{itemize}
\begin{table}[thbp]
	\centering
	\setlength{\tabcolsep}{12pt}
	\caption{Comparison between the structured attention composition module (SAC) and traditional attention module, measured by mAP (\%) on THUMOS14 dataset.}
	\begin{tabular}{l|cc}
    \toprule
    Setups & SSAD\cite{lin2017single} & G-TAD\cite{xu2020g} \\
    \midrule
    Baseline + Sequential Attention & 48.3  & 43.2 \\
    Baseline + non-local \cite{wang2018non} & 48.4  & 43.3 \\
    Baseline + Composed Attention & 48.8  & 43.7 \\
    Baseline + SAC & 51.4  & 44.4 \\
    \bottomrule
    \end{tabular}%
	\label{tabAblStructure}%
	\vspace{-0.5cm}
\end{table}
\begin{itemize}
    \item Non-local attention. Given a video segment, we can employ the non-local mechanism \cite{wang2018non} to connect a feature and all other features. It explicitly considers all temporal features to estimate the frame attention, but can only implicitly estimate the modality attention in a data-driven manner.
\end{itemize}
\noindent
As shown in Table \ref{tabAblStructure}, based on SSAD \cite{lin2017single}, the sequential attention mechanism achieves an mAP of 48.3\%, while the non-local mechanism \cite{wang2018non} gets an mAP of 48.4\%. The localization results are not accurate enough, because these mechanisms only integrate local and global information but do not thoroughly exploit structure cues. In contrast, the optimal transport process explicitly exploits the structure cue between the frame attention and the modality attention via the learnable frame-modality structure matrix, and brings adequate gains.

\begin{figure*}[thbp]
	\graphicspath{{figure/}}
	\centering
	\includegraphics[width=0.8\linewidth]{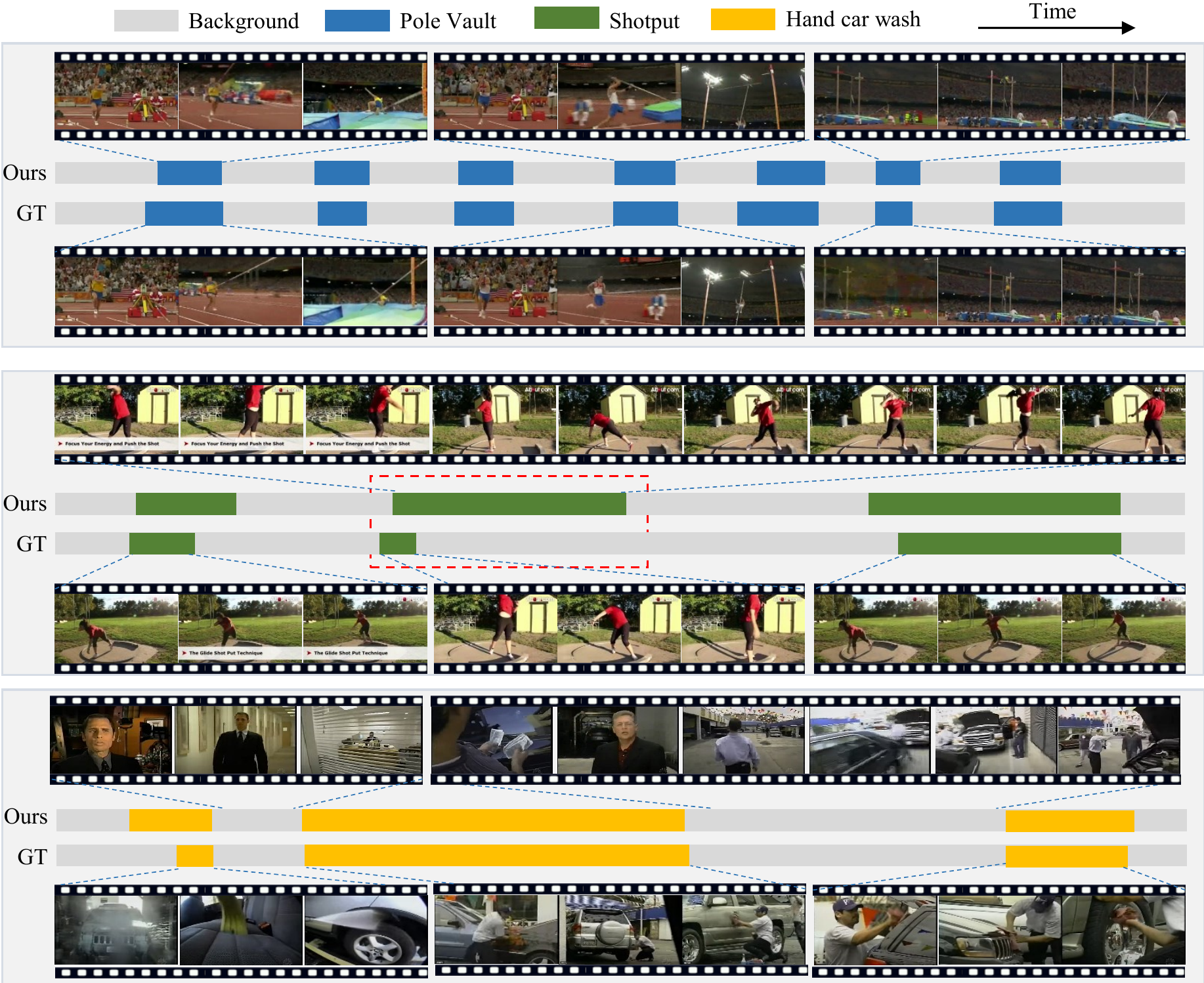}
	\caption{Visualization of the action localization results. In the first video, our method can precisely localize the \textit{Pole Vault} instances. In the second video, our method discovers two \textit{Shotput} instances but makes a mistake for an incomplete case (shown in red dotted box). As for ActivityNet v1.3, our method accurately localizes the \textit{Hand Car Wash} instance (the third video).}
	\label{figVisPosNeg}
\end{figure*}

\begin{table*}[htbp]
  \centering
  \caption{Computational cost analysis of the proposed structured attention composition module (SAC), in terms of parameter (M) and FLOPs (G).}
  \setlength{\tabcolsep}{4pt}
    \begin{tabular}{lllllllll}
    \toprule
    \multicolumn{1}{l|}{Method} & \multicolumn{1}{c}{SSAD\cite{lin2017single}} & \multicolumn{1}{c|}{SSAD\cite{lin2017single}+SAC} & \multicolumn{1}{c}{BMN\cite{lin2019bmn}} & \multicolumn{1}{c|}{BMN\cite{lin2019bmn}+SAC} & \multicolumn{1}{c}{G-TAD\cite{xu2020g}} & \multicolumn{1}{c|}{G-TAD\cite{xu2020g}+SAC} & \multicolumn{1}{c}{Liu \etal\cite{liu2021multi}} & \multicolumn{1}{c}{Liu \etal\cite{liu2021multi}+SAC} \\
    \midrule
    \multicolumn{1}{l|}{Parameters (M)} & \multicolumn{1}{c}{4.29} & \multicolumn{1}{c|}{5.08} & \multicolumn{1}{c}{4.98} & \multicolumn{1}{c|}{5.46} & \multicolumn{1}{c}{5.62} & \multicolumn{1}{c|}{6.37} & \multicolumn{1}{c}{18.03} & \multicolumn{1}{c}{18.95} \\
    \multicolumn{1}{l|}{FLOPs (G)} & \multicolumn{1}{c}{0.42} & \multicolumn{1}{c|}{0.62} & \multicolumn{1}{c}{91.2} & \multicolumn{1}{c|}{91.3} & \multicolumn{1}{c}{150.24} & \multicolumn{1}{c|}{150.78} & \multicolumn{1}{c}{14.06} & \multicolumn{1}{c}{14.58} \\
    \midrule
    \multicolumn{9}{l}{Because each method employs specific input datas (\eg different feature dimensions), the extra parameters brought by the SAC module are different.} \\
    \end{tabular}%
  \label{tabParaFlops}%
\end{table*}%

\begin{figure}[thbp]
     \centering
     \begin{subfigure}[b]{0.22\textwidth}
         \centering
         \setlength{\abovecaptionskip}{0.cm}
         \graphicspath{{figure/}}
         \includegraphics[width=\textwidth]{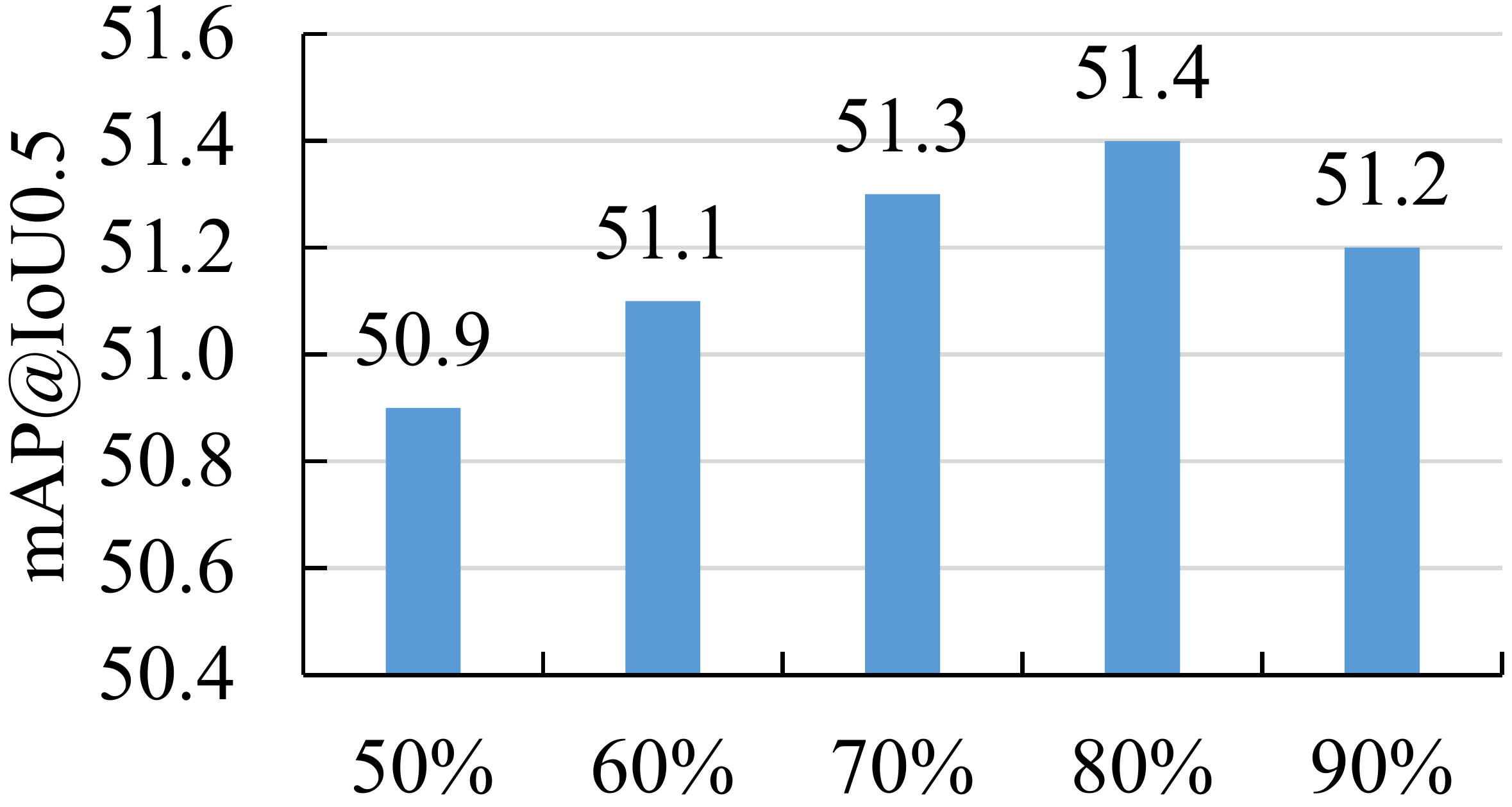}
         \caption{\footnotesize Ratio $\eta$ in the smoothness regularizer.}
         \label{figAblEta}
     \end{subfigure}
     \begin{subfigure}[b]{0.22\textwidth}
         \centering
         \setlength{\abovecaptionskip}{0.cm}
         \graphicspath{{figure/}}
         \includegraphics[width=\textwidth]{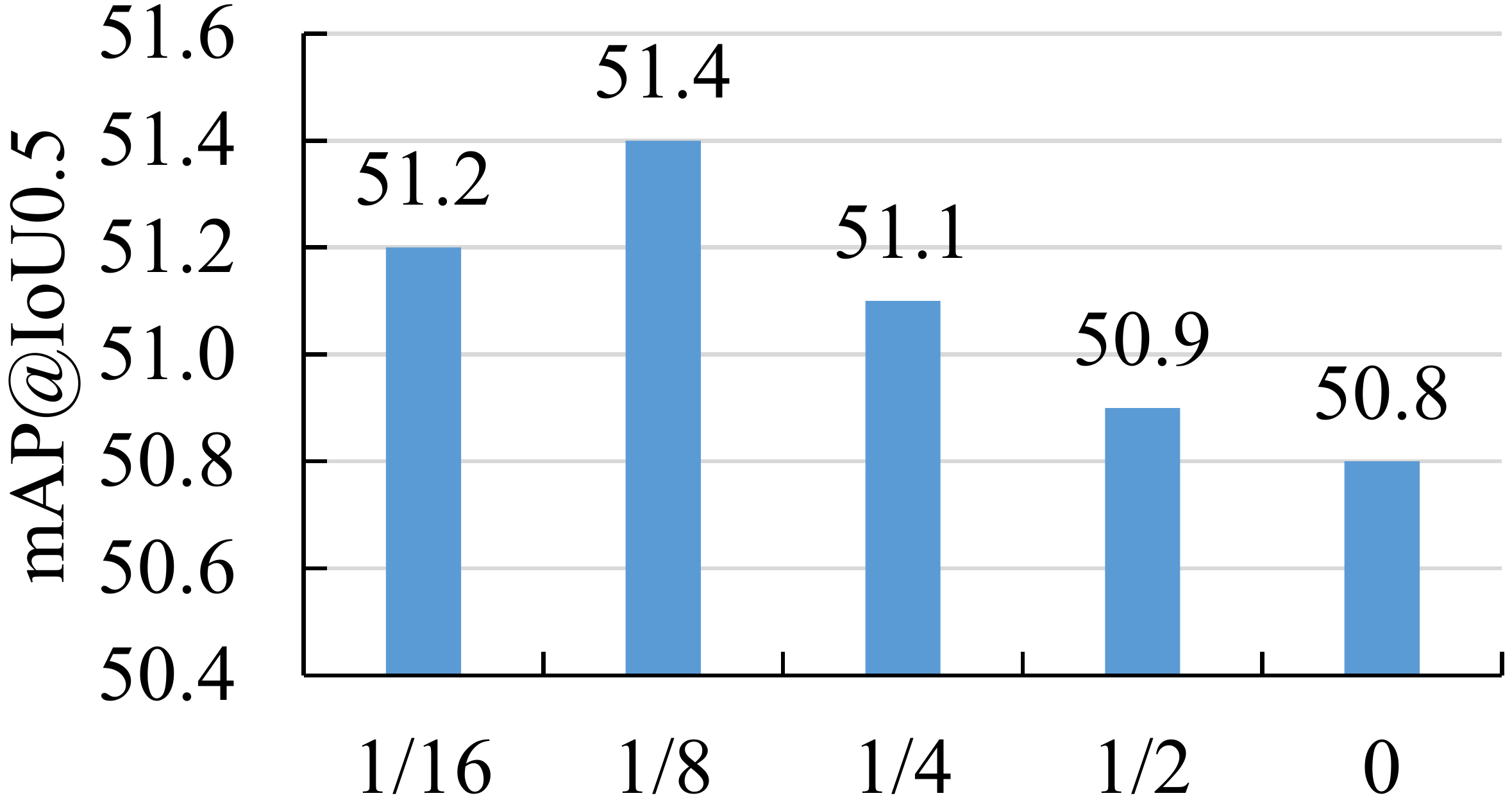}
         \caption{\footnotesize Ratio $k$ in the action-aware pooling procedure.}
         \label{figAblK}
     \end{subfigure}
     \begin{subfigure}[b]{0.22\textwidth}
         \centering
         \setlength{\abovecaptionskip}{0.cm}
         \graphicspath{{figure/}}
         \includegraphics[width=\textwidth]{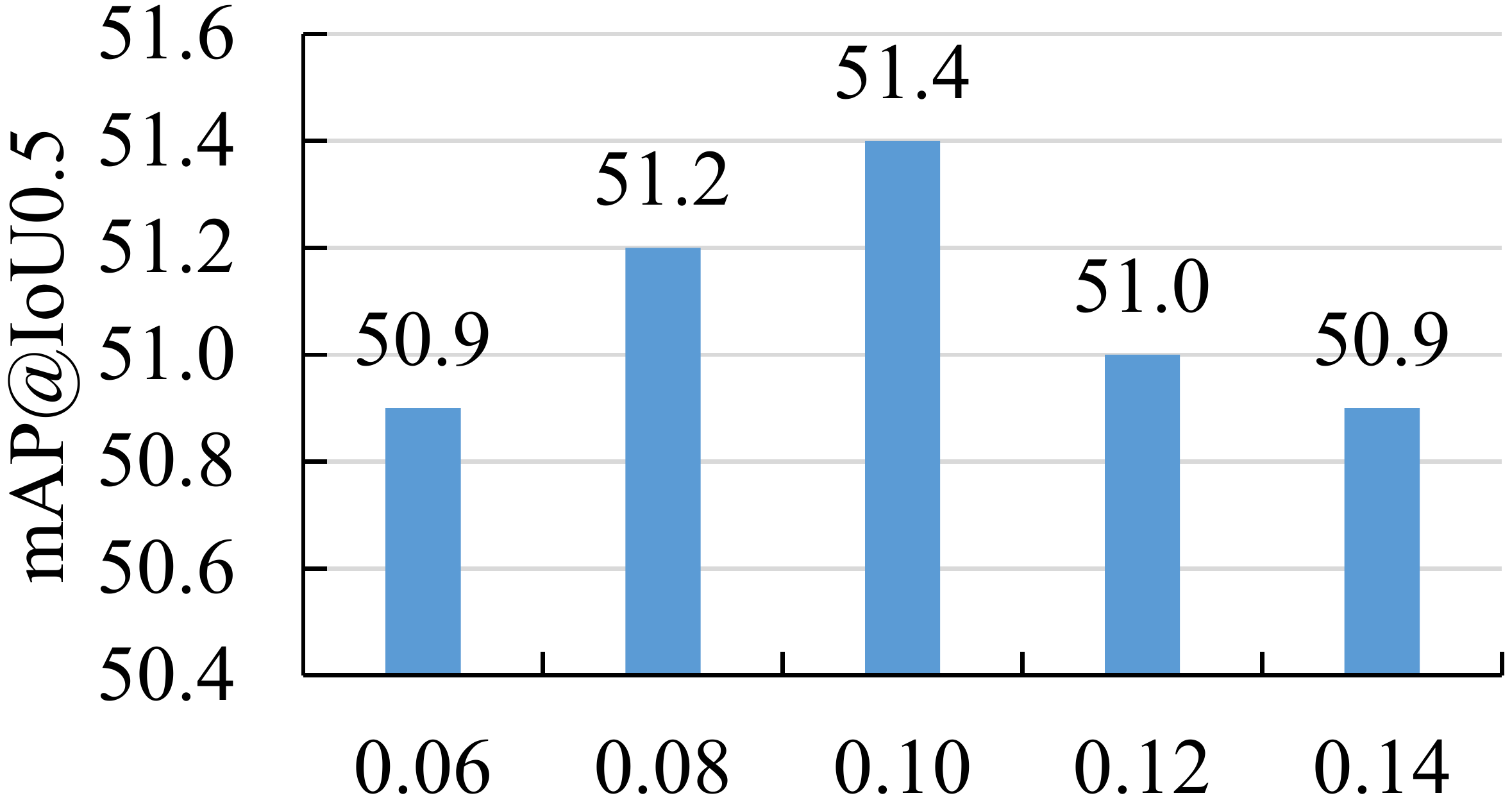}
         \caption{\footnotesize Coefficient $\lambda_{\rm s}$ for the smoothness loss $\mathcal{L}_{\rm s}$.}
         \label{figAblSmt}
     \end{subfigure}
     \begin{subfigure}[b]{0.22\textwidth}
         \centering
         \setlength{\abovecaptionskip}{0.cm}
         \graphicspath{{figure/}}
         \includegraphics[width=\textwidth]{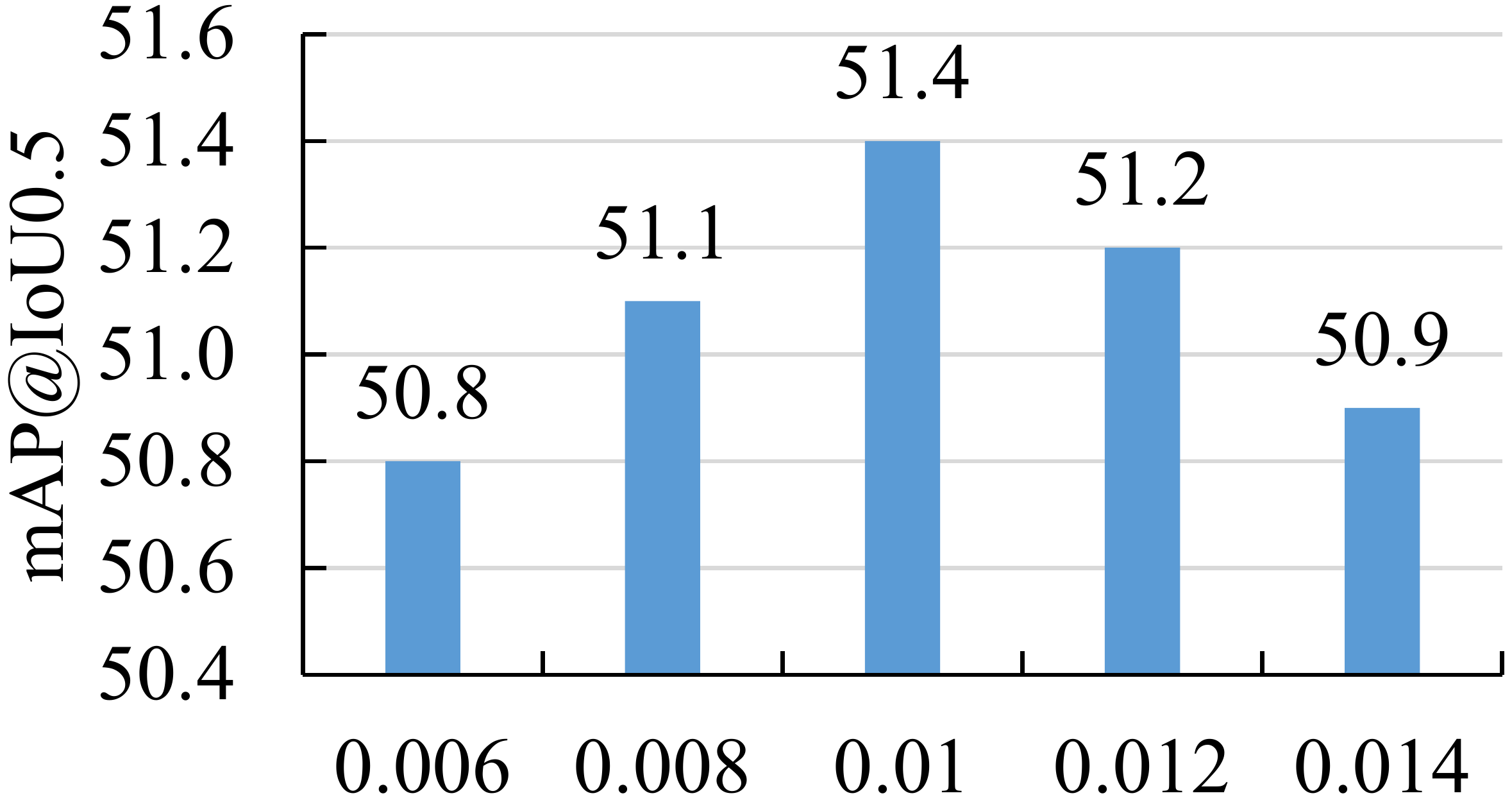}
         \caption{\footnotesize Coefficient $\lambda_{\rm F}$ for the normalization loss $\mathcal{L}_{\rm F}$.}
         \label{figAblF}
     \end{subfigure}
        \caption{Ablation studies about hyper-parameters in the proposed structured attention composition module, measured by mAP (\%) on THUMOS14 dataset.}
        \label{figAblation}
\end{figure}

\textbf{Hyper-parameters.} Fig. \ref{figAblation} reports ablation studies about hyper-parameters. In the smoothness regularizer, the hyper-parameter $\eta$ influences the number of selected minimal elements, where 80\% is a proper choice. In the action-aware pooling procedure, $k$ controls the ratio of confident action features. We found top $1/8$ features are beneficial to predict the modality attention, and redundant features may bring background noises. In the complete loss function (\ie Eq. (\ref{edLoss})), we study the coefficients $\lambda_{\rm s}$ and $\lambda_{\rm F}$ for the temporal smoothness loss $\mathcal{L}_{\rm s}$ and the F-norm loss $\mathcal{L}_{\rm F}$, respectively, and find the proper value as 0.10 and 0.01.

\textbf{Computational costs.} To further study the effectiveness of the structured attention composition module, Table \ref{tabParaFlops} reports parameters and FLOPs based on four different methods \cite{lin2017single, lin2019bmn, xu2020g, liu2021multi}. In particular, we employ the pytorch-OpCounter tool\footnote{\url{https://github.com/Lyken17/pytorch-OpCounter}} to measure the parameters of each method. As for FLOPs, considering each method employs specific input datas, \eg different temporal lengths and different feature dimensions, we keep the same input with the original method and set the batchsize as 1. As shown in Table \ref{tabParaFlops}, the proposed structured attention composition module only slightly increases parameters and FLOPs to the four existing methods \cite{lin2017single, lin2019bmn, xu2020g, liu2021multi}, demonstrating its effectiveness to the temporal action localization task.

\subsection{Qualitative visualization.}

Fig. \ref{figVisualization} qualitatively shows the effectiveness of the structured attention composition module. Considering the localization results of ``SSAD" and ``SSAD+SAC", we can find introducing structured attention composition into SSAD \cite{lin2017single} assists to precisely localize action boundaries. Besides, Fig. \ref{figVisualization} exhibits the effectiveness of frame-modality attention. As discussed in Fig. \ref{fig-motivation_RGB_flow_preference}, the instance \textit{Throw Discus} is confusing in the appearance modality and the instance \textit{Soccer Penalty} may be misled by the motion modality. In Fig. \ref{figVisualization}, the proposed structured attention composition module appropriately assigns attention weights to different modalities, \ie relying more on the motion modality for localizing the \textit{Throw Discus} instance and focusing on the appearance modality for discovering the \textit{Soccer Penalty} instance.

In addition, Fig. \ref{figVisPosNeg} qualitatively exhibits the localization results. On the one hand, our method can precisely localize most action instances, demonstrating its efficacy in tackling the temporal action localization task. On the other hand, the proposed method makes a mistake in the \textit{Shotput} video because of an incomplete case. Specifically, a coach is demonstrating the \textit{Shotput} action. She places the shot on her shoulder and rotates the body, but does not throw out the shot, which is quite different from regular instances. As a result, the incomplete segment is not annotated as the \textit{Shotput} instance. In the future, modeling the completeness of the action instance is a promising way to alleviate this issue.

\section{Conclusion}
\label{secConclusion}

In this paper, we make an early effort to study temporal action localization from the perspective of multi-modality feature learning. For this purpose, we propose a structured attention composition module to obtain the frame-modality attention, where the fitness between modality attention and frame attention is justified by an optimal transport process. Extensive experiments on four state-of-the-art methods demonstrate that our approach can serve as a plug-and-play module and consistently bring performance improvements. It should be noticed that we follow previous works \cite{long2019gaussian, 2020Revisiting, zeng2021graph, yang2021background} and extract video features in advance. However, in a practical scenario, the action localization algorithm should tackle raw videos, extract features and discover action instances, where existing feature extracting works\cite{carreira2017quo, zhao2017cuhk} with high computations would be inconvenient. In the future, it is a promising direction to improve the efficiency of the action localization algorithm, alternatively, to develop the feature coding strategy \cite{lin2020key} and transmit high computations to back-end computing center. Besides, we plan to introduce the structured attention composition module to other related tasks, \eg online action detection \cite{yang2022colar}, and further verify its effectiveness for the multi-modality feature learning.

\bibliographystyle{IEEEtran}
\bibliography{egbib}

\begin{IEEEbiography}[{\includegraphics[width=1in,height=1.25in,clip,keepaspectratio]{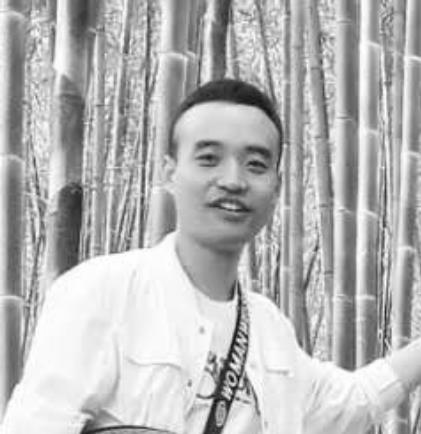}}]{Le Yang} received his B.E. degree from Northwestern Polytechnical University, Xi'an, China, in 2016. He is currently a Ph.D. candidate in the School of Automation at Northwestern Polytechnical University. His research interests include temporal action localization, video object segmentation and weakly supervised learning.
\end{IEEEbiography}

\begin{IEEEbiography}[{\includegraphics[width=1in,height=1.25in,clip,keepaspectratio]{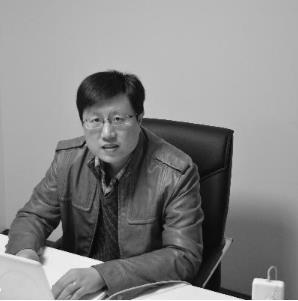}}]{Junwei Han}
	is currently a Professor in the School of Automation, Northwestern Polytechnical University. His research interests include computer vision, pattern recognition, remote sensing image analysis, and brain imaging analysis. He has published more than 70 papers in top journals such as IEEE TPAMI, TNNLS, IJCV, and more than 30 papers in top conferences such as CVPR, ICCV, MICCAI, and IJCAI. He is an Associate Editor for several journals such as IEEE TNNLS and IEEE TMM.
\end{IEEEbiography}

\begin{IEEEbiography}[{\includegraphics[width=1in,height=1.25in,clip,keepaspectratio]{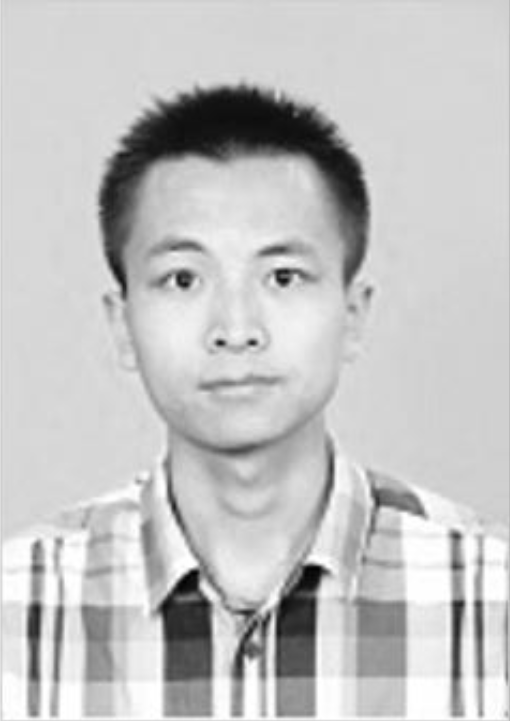}}]{Tao Zhao} received his M.S. degree from Northwestern Polytechnical University, Xi'an, China, in 2018. He is currently a Ph.D. candidate in the School of Automation at Northwestern Polytechnical University. His research interests include video temporal action localization and weakly supervised learning.
\end{IEEEbiography}

\begin{IEEEbiography}[{\includegraphics[width=1in,height=1.25in,clip,keepaspectratio]{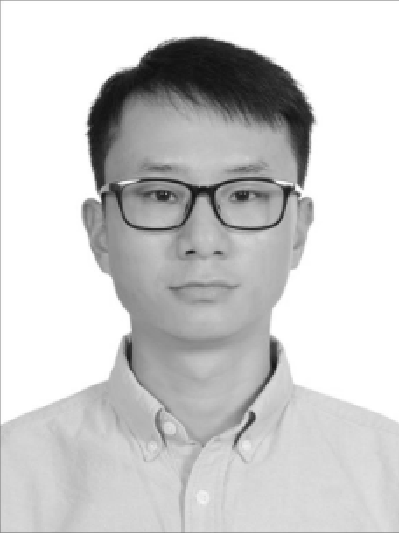}}]{Nian Liu}
	is currently a researcher with the Inception Institute of Artificial Intelligence, Abu Dhabi, UAE. He received the Ph.D. degree and the B.S. degree from the School of Automation at Northwestern Polytechnical University, in 2020 and 2012, respectively. His research interests include computer vision and machine learning, especially on saliency detection and deep learning.
\end{IEEEbiography}

\begin{IEEEbiography}[{\includegraphics[width=1in,height=1.25in,clip,keepaspectratio]{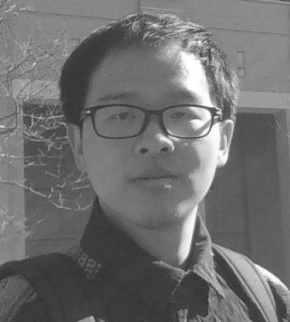}}]{Dingwen Zhang} received his Ph.D. degree from the Northwestern Polytechnical University, Xi'an, China, in 2018. He is currently a Professor in the School of Automation, Northwestern Polytechnical University. From 2015 to 2017, he was a visiting scholar at the Robotic Institute, Carnegie Mellon University. His research interests include computer vision and multimedia processing, especially on saliency detection, video object segmentation, temporal action localization and weakly supervised learning.
\end{IEEEbiography}

\end{document}